\documentclass[11pt]{article}

\PassOptionsToPackage{table}{xcolor}

\usepackage[final]{acl}

\usepackage{times}
\usepackage{latexsym}

\usepackage[T1]{fontenc}
\usepackage[utf8]{inputenc}

\usepackage{microtype}

\usepackage{inconsolata}

\usepackage{graphicx}

\usepackage[export]{adjustbox} % 在导言区添加
\usepackage{multirow}
\usepackage{booktabs}
\usepackage{tabularx}
\usepackage{arydshln}
\newcolumntype{Y}{>{\centering\arraybackslash}X}
\usepackage{amsmath,amsfonts}
\usepackage{cuted}  % 导言区添加
\usepackage{subcaption}
\definecolor{mycustomcolor}{HTML}{FFF2CD}
\usepackage[export]{adjustbox}
\usepackage{enumitem}
\usepackage{algorithm}
\usepackage{algpseudocode}
\title{SmartThinker: Learning High-Density Reasoning via Step-Level Token Allocation}

\author{
  \textbf{Xingyang He\textsuperscript{1,2}} \quad
  \textbf{Jie Liu\textsuperscript{1,$\dagger$}} \quad
  \textbf{Xiao Ling\textsuperscript{1}} \\
  \textsuperscript{1}National Key Laboratory of Intelligent Tracking and Forecasting for Infectious Diseases, \\
  Engineering Research Center of Trusted Behavior Intelligence, Ministry of Education, \\
  Tianjin Key Laboratory of Intelligent Robotic (tKILIR), \\
  Institute of Robotics and Automatic Information System (IRAIS), Nankai University \\
  \textsuperscript{2}Shanghai Innovation Institute \\
  \texttt{\{xingyanghe,lingxiao\}@mail.nankai.edu.cn}, \texttt{jliu@nankai.edu.cn}}

\begin{document}
\maketitle
{\renewcommand{\thefootnote}{$\dagger$}\footnotetext{Corresponding author.}}
\begin{abstract}

Efficient reasoning is critical for large reasoning models (LRMs) deployed in agent systems with strict context constraints. Existing methods predominantly rely on global length penalties that overlook step-level structure, inadvertently compressing critical derivations while retaining low-value detours. Hence, we argue for a principled, step-aware alternative that preserves critical steps and compresses low-value detours. However, it raises three fundamental challenges: the difficulty of obtaining step-level supervision, the need for reward signals that induce differentiated length pressure, and the requirement to balance training stability with exploration on complex problems. We propose SmartThinker, a step-level token allocation framework that learns to selectively preserve or compress individual reasoning steps. At its core is Step-Level Length Control Policy Optimization (SCPO), which provides label-free feedback by online estimation of each step's contribution from the change in correct-answer likelihood after removing it. We further introduce reasoning density, defined as contribution per token, to relax length pressure on high-value steps while intensifying it on low-density ones. Finally, SCPO stabilizes training through length-bias-corrected advantages and difficulty-adaptive clipping, achieving superior accuracy-efficiency trade-offs across nine in-domain and out-of-domain benchmarks.

% We conduct extensive experiments across nine benchmarks. The results show that SmartThinker reduces redundant reasoning while improving accuracy over efficient-reasoning baselines. 

% Across mathematical reasoning benchmarks and backbone scales, SmartThinker reduces redundant reasoning while outperforming efficient-reasoning baselines. These results indicate that LRMs can learn more direct reasoning paths, freeing context for denser inference in agentic, budget-constrained settings.
\end{abstract}

\begin{figure}[t]
    \centering
    \includegraphics[width=\linewidth]{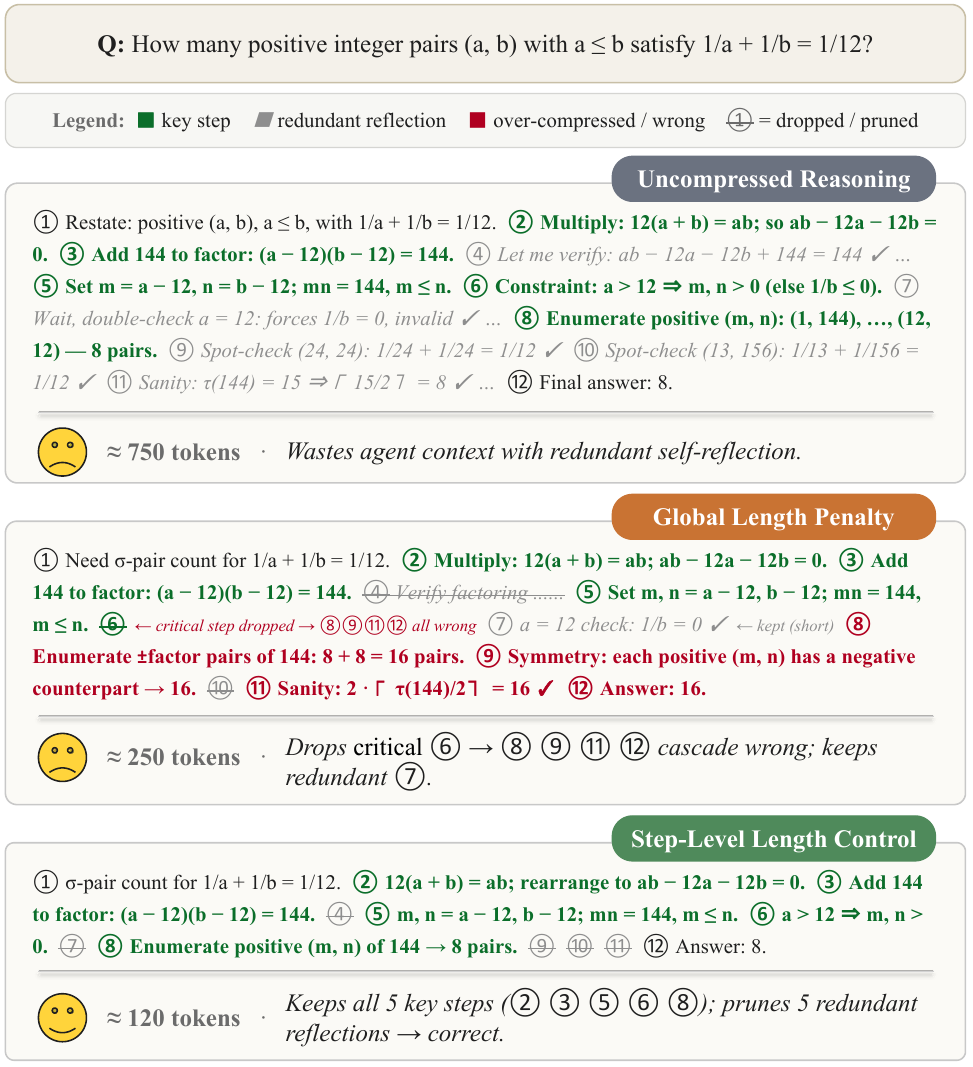}
    \caption{Comparison of reasoning traces under different length-control settings. Uncompressed reasoning wastes context on redundant detours; global length penalties shorten responses uniformly and can remove critical derivations; Step-level length control compresses low-value steps while preserving critical ones.}
    \label{fig:motivating_case}
\end{figure}

\section{Introduction}

Efficient reasoning is critical for deploying large reasoning models (LRMs) in agentic frameworks. Although longer reasoning trajectories can improve complex problem solving \cite{guo2025deepseek,jaech2024openai,yang2024qwen2,luo2024improve}, they often include redundant verification and repetitive reflection \cite{ye2025limo, kumar2025overthink,fu2024efficiently,sui2025stop,qu2025survey}, reducing agent throughput and crowding out contextual evidence needed for downstream decisions. Prior works primarily improve reasoning efficiency through global output shortening, either by imposing global length penalties during reinforcement learning (RL) \cite{aggarwal2025l1,arora2025training,hou2025thinkprune,zhang2025grpo} or by routing queries to short- or long-reasoning modes based on estimated difficulty \cite{fang2025thinkless,zhang2025adaptthink,lou2025adacot,luo2025ada}.

Despite their effectiveness, global-level methods treat all reasoning steps as equally compressible and ignore step-level differences, which leads to compressing critical derivations while preserving low-value ones. Figure~\ref{fig:motivating_case} gives an intuitive example. Without compression, the model spends many tokens on redundant reasoning that consumes context. A global length penalty reduces the overall chain length, but may remove critical derivations together with redundant text. In contrast, step-level length control can preserve the steps that support the final answer while compressing those with little contribution. However, step-level optimization still faces three key challenges. First, step-level supervision is difficult to obtain because human annotations are expensive and external reward models introduce additional training cost and bias. Second, even if a step's contribution can be estimated, it must be converted into differentiated length pressure so that useful steps are protected and redundant ones are compressed. Third, training must remain stable while preserving exploration, since hard problems often require longer verification and correction.

To address these challenges, we propose SmartThinker, a step-level token allocation method that learns which reasoning steps should be preserved or compressed. SmartThinker contains two stages. The first stage uses rejection sampling and supervised fine-tuning (SFT) to warm up the model into a concise reasoning regime, reducing rollout cost and providing a stable initialization for later RL. The second stage introduces Step-Level Length Control Policy Optimization (SCPO), which performs fine-grained length control over individual reasoning steps. To provide label-free process feedback, SCPO estimates each step's answer contribution online by measuring how removing that step changes the likelihood of the correct answer. We further define reasoning density as answer contribution per generated token and uses it to guide the model to impose stronger penalties on low-density steps while relaxing constraints on high-density ones. To stabilize optimization, SCPO further applies length-bias-corrected advantage and difficulty-adaptive clipping, allowing the model to preserve stable reasoning paths on easier questions while maintaining exploration on harder ones.

% To address this limitation, we propose SmartThinker, a step-level token allocation method that learns where reasoning length should be preserved or compressed. 
% %Instead of applying the same length pressure to the whole response, SmartThinker uses step-level answer contribution to protect useful derivations and impose stronger compression on low-density steps.
% Our method comprises two stages: a supervised warm-up phase followed by a RL phase. The warm-up phase uses rejection sampling and supervised fine-tuning (SFT) to quickly adapt a reasoning model to a concise regime. Although this process may cause a slight performance drop, it significantly accelerates convergence in the subsequent RL phase. The RL phase introduces Step-Level Length Control Policy Optimization (SCPO) to perform fine-grained length control across individual reasoning steps. Instead of applying the same length pressure to the whole response, SCPO estimates each step's answer contribution online without human process labels or an external reward model and uses this estimate to construct density-aware step rewards. These rewards guide the model to impose stronger penalties on low-density steps while relaxing constraints on high-density ones. In addition,
% SCPO further introduces length-bias-corrected credit assignment via S-GAE and difficulty-adaptive policy updates, allowing the model to preserve stable reasoning paths on easier questions while maintaining exploration on harder ones. 

We evaluate SmartThinker across nine in-domain and out-of-domain benchmarks. The results show that SmartThinker reduces redundant reasoning while outperforming efficient-reasoning baselines. In context-budgeted settings, the saved tokens translate into more useful reasoning opportunities, supporting the practical value of step-level token allocation.

In summary, our contributions are as follows:
\begin{itemize}[noitemsep]
    \item We introduce the perspective of reasoning density, reformulating efficient reasoning as a step-level token allocation problem and estimating each step’s answer contribution through its effect on the probability of the correct answer.

    \item  We propose SmartThinker, a two-stage framework for learning high-density reasoning flows, where SCPO enables fine-grained step-level length control without human process labels or external reward models.
    % We propose SmartThinker and SCPO, which convert online estimates of step-level answer contribution into density-aware rewards, length-bias-corrected credit assignment and difficulty-adaptive policy updates without human process labels or an external reward model.

    \item We systematically evaluate SmartThinker across in-domain, context-budgeted, out-of-domain, and ablation settings, showing that it reduces redundant reasoning while improving the accuracy-efficiency trade-off.
\end{itemize}

\begin{figure*}[ht]
    \begin{center}
		\includegraphics[width=0.9\linewidth]{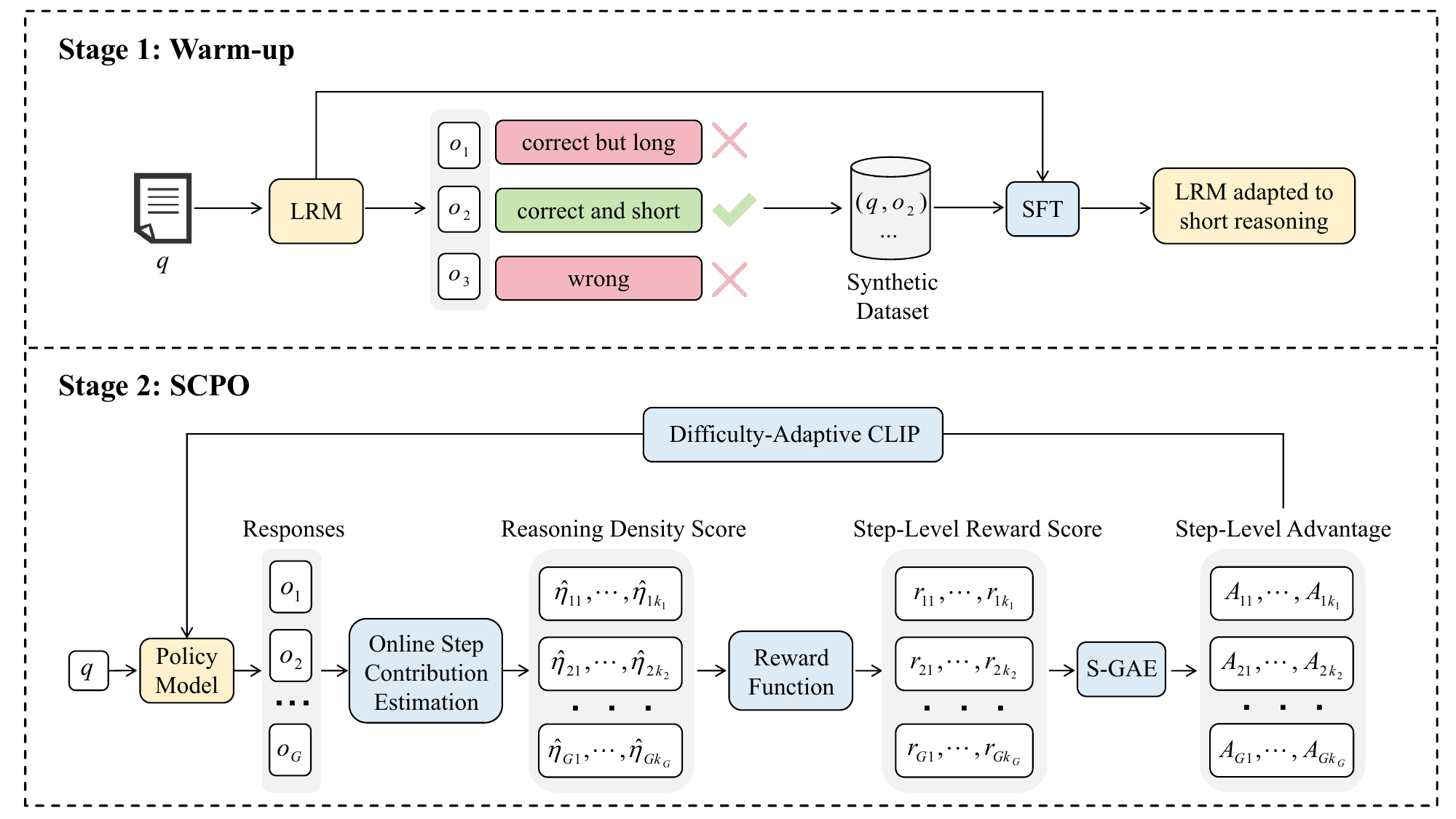}
    \end{center}
    \caption{Overview of the proposed SmartThinker.}
    \label{SmartThinker}
\end{figure*}

\section{Related Work}
\vspace{-0.1em}
\paragraph{Efficient and Adaptive Reasoning.}
% Long reasoning traces improve the problem-solving ability of LRMs, but often introduce redundancy and serving cost \cite{ye2025limo,kumar2025overthink,fu2024efficiently,sui2025stop,qu2025survey}. Prior work reduces this cost through concise reasoning data or budget-conditioned training \cite{xia2025tokenskip,han2024token,xu2025chain,zhang2025lightthinker,yan2025inftythink}, global length penalties in RL \cite{aggarwal2025l1,arora2025training,hou2025thinkprune,zhang2025grpo}, and adaptive short/long reasoning modes \cite{fang2025thinkless,zhang2025adaptthink,lou2025adacot,luo2025ada}. These methods improve efficiency, but their control signal is defined over the whole response or reasoning mode. SmartThinker is complementary: it controls token allocation within a reasoning trajectory, preserving high-contribution steps while compressing low-density detours.

Long reasoning traces can improve the problem-solving ability of LRMs, but they also introduce substantial redundancy and serving cost \cite{ye2025limo,kumar2025overthink,fu2024efficiently,sui2025stop,qu2025survey}. Existing work on efficient reasoning typically controls this cost at the global level. One line of work constructs concise reasoning data or trains models to write shorter intermediate derivations under a target budget \cite{xia2025tokenskip,han2024token,xu2025chain,zhang2025lightthinker,yan2025inftythink}. Another line incorporates global length penalties into RL objectives, encouraging the model to reduce the total number of generated tokens \cite{aggarwal2025l1,arora2025training,hou2025thinkprune,zhang2025grpo,wu2025lapo,huang2025hapo,liu2025alrs,tu2025when}. A third line learns adaptive short/long reasoning behavior, either by selecting a reasoning mode or by triggering chain-of-thought only when the problem appears sufficiently difficult \cite{fang2025thinkless,zhang2025adaptthink,lou2025adacot,luo2025ada}. These methods improve efficiency, but their primary control signal is still defined over the whole response or the whole reasoning mode. SmartThinker is complementary: it focuses on how tokens are allocated within a reasoning trajectory, so that high-contribution steps can be preserved while low-contribution steps are selectively compressed.

\paragraph{Process Supervision and Credit Assignment.}
Process supervision assigns feedback to intermediate steps rather than only to final answers \cite{lightman2023let,luo2024improve}, making it closely related to credit assignment in long-chain reasoning. Existing methods often depend on human step labels, model-generated annotations, or a separate process reward model. SCPO instead estimates each step's answer contribution online by measuring how removing the step changes the probability of the correct answer. It then converts this estimate into step-level length-control rewards, linking process feedback to token efficiency: a step should be preserved only when its answer-relevant contribution justifies its token cost. Concurrent work also moves control signals below the trajectory level, including step-level advantage selection for stabilizing efficient reasoning \cite{wang2025sas}, stepwise length penalization \cite{li2026stepwise}, and step-level credit assignment for RL training \cite{li2026ssvpo}. SCPO differs in that the step score is an online, contribution-aware estimate of how much a step helps the final answer, and it is used to allocate tokens selectively within a single trajectory rather than to penalize steps or select steps globally.

\paragraph{Agent Harness and Context-Efficient Learning.}
Reasoning models are increasingly deployed in agentic systems, where they interact with tools, search procedures, retrieved memories, and environment feedback \cite{wu2025agentic,chen2025learning}. In such settings, reasoning tokens compete with observations, intermediate state, conversation history, and candidate plans within the same context window. Token efficiency is therefore not only a serving concern; it affects how much useful state an agent can retain while planning and acting. SmartThinker addresses this pressure at training time by encouraging denser reasoning trajectories, leaving more context budget for later tool calls, memory updates, and multi-branch search.

\section{Method}
\vspace{-0.2em}
SmartThinker learns high-density reasoning by dense process reward signals. As illustrated in Figure~\ref{SmartThinker}, it first uses a short-reasoning warm-up for stable initialization, and then applies Step-Level Length Control Policy Optimization (SCPO). SCPO does not optimize for uniformly shorter outputs; it learns to spend tokens on reasoning steps with high answer contribution per token.
Algorithm~\ref{alg:scpo} provides a compact overview of the SCPO update, and the following subsections unpack its estimation, reward, credit assignment, and clipping components.

\subsection{Short-Reasoning Warm-up}
Direct RL from a long-reasoning model is inefficient because early rollouts are often lengthy and low-density. We therefore use rejection sampling followed by supervised fine-tuning (SFT) as an initialization step. For each training question, the current model generates multiple candidate responses; we retain the shortest correct response and fine-tune on the resulting synthetic data. Since the data come from the model itself, the warm-up stays close to the model distribution while moving the policy into a concise regime. Fine-grained length control is learned later by SCPO.

\begin{algorithm}[t]
    \caption{Step-Level Length Control Policy Optimization}
    \label{alg:scpo}
    \begin{algorithmic}[1]
    \Require Question $q$, answer $a$, policy $\pi_{\theta}$, old policy $\pi_{\theta_{\mathrm{old}}}$, rollout size $G$
    \State Sample $G$ responses $\{o_i\}_{i=1}^{G}$ from $\pi_{\theta_{\mathrm{old}}}$
    \State Split each $o_i$ into steps $S_i=\{s_{i,1},\ldots,s_{i,n_i}\}$
    \State Compute answer correctness and difficulty $\rho(q)=1-N_c/G$
    \For{each step $s_{i,j}$}
        \State Estimate raw contribution $m_{i,j}$ by removing $s_{i,j}$
        \State Compute density $\eta_{i,j}$ and normalized allocation score $\widehat{\eta}_{i,j}$
        \State Compute the step reward $r_{i,j}$ with density- and difficulty-aware penalties
    \EndFor
    \State Normalize step rewards to obtain $\widetilde{r}_{i,j}$
    \State Compute $A_{i,j}^{\mathrm{S\text{-}GAE}}=\sum_{m=0}^{n_i-j}\gamma^m \widetilde{r}_{i,j+m}$
    \State Assign $A_{i,j}^{\mathrm{S\text{-}GAE}}$ to all tokens in step $s_{i,j}$
    \State Set adaptive clipping bounds $\varepsilon_{\mathrm{low}}$ and $\varepsilon_{\mathrm{high}}$
    \State Update $\pi_{\theta}$ with the clipped SCPO objective using bounds $[1-\varepsilon_{\mathrm{low}}, 1+\varepsilon_{\mathrm{high}}]$
    \end{algorithmic}
\end{algorithm}

\subsection{Online Step Contribution Estimation}
For each training pair $(q,a)$, SCPO samples $G$ responses from the old policy. Each response $o_i$ contains a reasoning trajectory and a final predicted answer:
\begin{align}
o_i = (S_i, a_i), \quad
S_i=\{s_{i,1}, s_{i,2}, \ldots, s_{i,n_i}\},
\end{align}
where $s_{i,j}$ is the $j$-th reasoning step, $l_{i,j}$ is its token length, and $n_i$ is the number of steps. SCPO uses this step structure to preserve useful reasoning while compressing low-value detours.

The sampled group also provides an online estimate of question difficulty:
\begin{align}
\rho(q)=1-\frac{N_c}{G},
\label{eq:problem_difficulty}
\end{align}
where $N_c$ is the number of correct responses. A larger $\rho(q)$ indicates a harder question for the current policy, and is used later to control both step-count penalty and exploration.

SCPO estimates each step's contribution online without human process labels or an external reward model. For a sampled trajectory $S_i$, let
\begin{align}
p_1 &= P_{\theta}(a \mid q, S_i), \\
p_2 &= P_{\theta}(a \mid q, S_i \setminus \{s_{i,j}\}),
\end{align}
where $p_1$ is the probability of the correct answer given the full trajectory, and $p_2$ is the probability after removing step $s_{i,j}$. We define the raw contribution score as
\begin{align}
m_{i,j} =
\begin{cases}
0, & \text{if } p_1 \le p_2, \\
\displaystyle \frac{p_1^2-p_2^2}{p_1^2}, & \text{if } p_1 > p_2 .
\end{cases}
\label{eq:raw_contribution}
\end{align}
The removal-based probability drop estimates the answer contribution of a step on answer correctness, while the squared difference sharpens the contrast between critical and ordinary steps. To account for token cost, we define reasoning density as contribution per token:
\begin{align}
\eta_{i,j}=\frac{m_{i,j}}{l_{i,j}},
\label{eq:reasoning_density}
\end{align}
Thus, a step receives relaxed length pressure only when its contribution is high relative to its length.

We add a lightweight safeguard for reflective steps on difficult questions. Steps containing transition markers such as ``but'' or ``however'' often indicate verification or correction. To avoid suppressing such behavior too aggressively, we define
\begin{align}
\widetilde{\eta}_{i,j}=\eta_{i,j}+\rho(q)I(s_{i,j}),
\end{align}
where
\begin{align}
I(s_{i,j}) =
\begin{cases}
\max_m \eta_{i,m}, & \text{if marker present},\\
0, & \text{otherwise}.
\end{cases}
\end{align}
This term protects verification behavior only when the rollout set suggests that the question is difficult. Finally, we min-max normalize $\widetilde{\eta}_{i,j}$ within the same correctness group:
\begin{align}
\widehat{\eta}_{i,j}
=
\frac{\widetilde{\eta}_{i,j}-\min(\mathcal{E}_{y_i})}
{\max(\mathcal{E}_{y_i})-\min(\mathcal{E}_{y_i})+\epsilon},
\end{align}
where $y_i=\mathbf{1}[a_i=a]$ and $\mathcal{E}_{y_i}$ denotes the set of density-based scores with the same correctness label as response $i$. The resulting $\widehat{\eta}_{i,j}$ is the allocation score used by the reward.

\subsection{Step-Level Reward as Token Allocation}
SCPO rewards correct responses while penalizing low-density length. The reward for step $s_{i,j}$ is
\begin{align}
r_{i,j}
&=
\mathbf{1}[a_i=a]\,
\bigl(1-k_1\sigma(\widetilde{l}_{i,j})\bigr) \notag\\
&\quad \cdot
\bigl(1-k_2\sigma(\widetilde{n}_{i})\bigr).
\end{align}
Here, $\sigma(\cdot)$ is the sigmoid function and
\begin{align}
k_1 &= k_0(1-\widehat{\eta}_{i,j}),\\
k_2 &= k_0(1-\rho(q)).
\end{align}
Here, $k_0$ is the base penalty coefficient. The density-aware coefficient $k_1$ gives low-density steps stronger length pressure, while the difficulty-aware coefficient $k_2$ penalizes excessive step counts more strongly on easier questions.

The normalized length and step-count terms are computed over correct responses:
\begin{align}
\widetilde{l}_{i,j}
=
\frac{l_{i,j}-\mathrm{mean}_{\{i,j:a_i=a\}}(l_{i,j})}
{\mathrm{std}_{\{i,j:a_i=a\}}(l_{i,j})},
\end{align}
\begin{align}
\widetilde{n}_{i}
=
\frac{n_i-\mathrm{mean}_{\{i:a_i=a\}}(n_i)}
{\mathrm{std}_{\{i:a_i=a\}}(n_i)}.
\end{align}
Thus, SCPO avoids rewarding short but wrong trajectories, and converts compression into step-level token allocation rather than global shortening.

\subsection{S-GAE for Step-Level Credit Assignment}
Naively accumulating future step rewards gives longer correct responses larger advantages simply because they contain more rewarded steps \cite{shao2024deepseekmath}. SCPO avoids this length bias with Step-Level Generalized Advantage Estimation (S-GAE).

We first normalize step rewards across the sampled group:
\begin{align}
\widetilde{r}_{i,j}
=
\frac{r_{i,j}-\mathrm{mean}_{i,j}(r_{i,j})}
{\mathrm{std}_{i,j}(r_{i,j})}.
\end{align}
The S-GAE advantage for step $s_{i,j}$ is then
\begin{align}
A_{i,j}^{\mathrm{S\text{-}GAE}}
=
\sum_{m=0}^{n_i-j}\gamma^m \widetilde{r}_{i,j+m},
\end{align}
where $\gamma\in(0,1]$ is a discount factor. The resulting advantage is assigned to all tokens in the corresponding step, providing dense feedback while keeping the step as the unit of credit assignment.

\subsection{Difficulty-Adaptive Clipping as Exploration Control}
Clip-Higher in DAPO~\cite{yu2025dapo} expands the upper clipping bound to mitigate entropy collapse and give low-probability tokens more room for exploration. This is useful for difficult problems, but it can be harmful for easy ones: when the model already follows a reliable reasoning path, allowing large probability increases for low-probability tokens may push it away from the correct solution. SCPO therefore keeps the clipped policy update but adapts the clipping range to the online difficulty estimate:
\begin{align}
\varepsilon_{\mathrm{low}}&=\varepsilon-\delta_1(1-\rho(q)), \label{down}\\
\varepsilon_{\mathrm{high}}&=\varepsilon+\delta_2\rho(q), \label{up}
\end{align}
where $\delta_1$ and $\delta_2$ control the adjustment magnitude. Easy questions receive tighter updates to preserve known correct paths, while difficult questions receive a larger upper bound to explore promising low-probability tokens.

\section{Experiments}

\subsection{Models and Datasets}
DeepSeek-R1-Distill-Qwen-1.5B, DeepScaleR-1.5B-Preview, and DeepSeek-R1-Distill-Qwen-7B are used as base models. We train these models on a randomly sampled subset of 3,000 QA pairs from the DeepScaleR-Preview dataset \cite{luo2025deepscaler}.
For evaluation, we use six in-domain benchmarks: AIME24, AIME25 \cite{veeraboina1983aime}, AMC23, MATH \cite{lewkowycz2022solving}, MinervaMATH \cite{hendrycks2021measuring}, Olympiad-Bench \cite{he2024olympiadbench} and three out-of-domain benchmarks: Live-Code-Bench \cite{jain2024livecodebench}, GPQA-Diamond \cite{rein2023gpqa}, and MuSR-MurderMysteries (MuSR-MM) \cite{sprague2024musr}.
Please see Appendix \ref{benchmark datasets} for more details.

% We primarily evaluate performance using three key metrics:
% (i) Pass@k: The probability that at least one of the top $k$ samples is correct under a given decoding strategy. This measures the upper bound of the model’s reasoning capability.
% (ii) Maj@k: The accuracy achieved through majority voting across $k$ samples. This metric assesses the stability of the model output in producing correct answers.
% (iii) AvgLen: The average length of all sampled responses, which serves as an indicator of the model’s reasoning efficiency.
% For AIME24, AIME25 and AMC23, we set $k = 10$ due to their relatively small number of test examples (fewer than 40). For the remaining datasets, we set $k = 5$.

We evaluate performance using three main metrics:
(i) Pass@k, the probability that at least one of the top $k$ samples is correct under a given decoding strategy, reflecting the upper bound of the model’s reasoning ability.
(ii) Maj@k, the accuracy obtained through majority voting across $k$ samples, which measures the stability of the model’s output.
(iii) AvgLen, the average length of all sampled responses, indicating the model’s reasoning efficiency.
For AIME24, AIME25, and AMC23, we set $k = 10$ due to the small size of their test sets. For all other datasets, we use $k = 5$.

% Note that we do not adopt Pass@1 as our primary evaluation metric because improvements in Pass@1 mainly reflect enhancements in output ranking \cite{zhu2025surprising}. In other words, the model becomes better at placing the correct answer first, rather than solving a greater number of previously unsolvable problems. In contrast, Pass@k (for $k>1$) allows the model to make multiple attempts, providing opportunities to explore potentially correct answers even if the first attempt is incorrect. This setting better reflects the model's true reasoning ability when allowed sufficient exploration.

Note that we do not use Pass@1 as our primary evaluation metric, since improvements in Pass@1 mainly reflect better output ranking \cite{zhu2025surprising}. In other words, the model becomes more skilled at placing the correct answer first rather than solving previously unsolvable problems. In contrast, Pass@k (for $k > 1$) allows multiple attempts, allowing the model to explore potentially correct answers even if the first attempt fails. This setting better reflects the model's true reasoning ability when sufficient exploration is allowed.

\subsection{Baselines}
% 我们选择的baseline包含两大类：（1）基于整体长度惩罚的模型，包括Alpha-0.1, L1-Max.(2) 长短推理混合的模型，包括Thinkless， AdaptThink.
The baselines we select can be grouped into two categories: (i) models based on global length penalty, including Alpha-0.1 \cite{arora2025training} and L1-Max \cite{aggarwal2025l1}, and (ii) models that incorporate both long- and short-reasoning modes, such as Thinkless \cite{fang2025thinkless} and AdaptThink \cite{zhang2025adaptthink}.

\subsection{Experiment Settings}

The hyperparameters are set to $k_{0} = 0.6$, $\gamma = 0.95$, $\delta_{1} = 0.03$, and $\delta_{2} = 0.08$. Detailed experimental settings are provided in Appendix \ref{setting details}.

\subsection{Main Results}
\begin{table}[t]
\renewcommand{\arraystretch}{1.05}
\setlength{\tabcolsep}{1pt}
\centering
\footnotesize
\caption{Average in-domain accuracy-efficiency trade-off. Averages are computed over the six in-domain benchmarks. Red. denotes token reduction relative to the corresponding original backbone; Dens. denotes average Pass@k per 1K generated tokens.}
\label{compact_main_results}
\begin{tabular}{@{}l*{5}{c}@{}}
\toprule
Models & Pass@k & Maj@k & AvgLen. & Red. & Dens. \\
\midrule
DeepSeek-R1-1.5B & 61.88 & 51.77 & 5768 & 0.0 & 10.7 \\
Alpha-0.1-1.5B & 65.17 & 53.08 & 4406 & 23.6 & 14.8 \\
Thinkless-1.5B & 63.69 & 51.93 & 4362 & 24.4 & 14.6 \\
\rowcolor{mycustomcolor}
SmartThinker-1.5B & \textbf{65.75} & \textbf{53.69} & \textbf{3651} & \textbf{36.7} & \textbf{18.0} \\
\midrule
DeepScaleR-1.5B & 66.53 & \textbf{57.51} & 4988 & 0.0 & 13.3 \\
L1-Max-1.5B & 64.33 & 53.98 & 2841 & 43.0 & 22.6 \\
\rowcolor{mycustomcolor}
SmartThinker-ScaleR & \textbf{66.65} & 55.38 & \textbf{2370} & \textbf{52.5} & \textbf{28.1} \\
\midrule
DeepSeek-R1-7B & 71.51 & 63.08 & 4987 & 0.0 & 14.3 \\
Alpha-0.1-7B & 73.83 & 65.10 & 4061 & 18.6 & 18.2 \\
AdaptThink-7B & 73.40 & 65.04 & 3833 & 23.1 & 19.1 \\
\rowcolor{mycustomcolor}
SmartThinker-7B & \textbf{75.26} & \textbf{66.59} & \textbf{3367} & \textbf{32.5} & \textbf{22.4} \\
\bottomrule
\end{tabular}
\end{table}

\begin{table}[t]
\renewcommand{\arraystretch}{1.05}
\setlength{\tabcolsep}{2pt}
\centering
\footnotesize
\caption{Context-budgeted AIME24 evaluation. Each entry reports the average number of correctly solved problems before the cumulative chat context reaches budget \(B\).}
\label{budget_main_results}
\begin{tabularx}{\columnwidth}{@{}lYYYY@{}}
\toprule
Models & 4K & 8K & 16K & 32K \\
\midrule
DeepSeek-R1-7B & 0.20 & 0.80 & 1.60 & 4.20 \\
Alpha-0.1-7B & 0.20 & 0.80 & 3.20 & 3.60 \\
AdaptThink-7B & \textbf{0.60} & 0.80 & 2.80 & 3.40 \\
\rowcolor{mycustomcolor}
SmartThinker-7B & \textbf{0.60} & 0.80 & \textbf{3.40} & \textbf{5.60} \\
\bottomrule
\end{tabularx}
\end{table}

\begin{table}[t]
\renewcommand{\arraystretch}{1.05}
\setlength{\tabcolsep}{2pt}
\centering
\footnotesize
\caption{Out-of-domain Pass@1 accuracy (\%). LCB, GPQA, and MuSR denote Live-Code-Bench, GPQA-Diamond, and MuSR-MM, respectively.}
\label{ood_main_results}
\begin{tabularx}{\columnwidth}{@{}lYYYY@{}}
\toprule
Models & LCB & GPQA & MuSR & Avg. \\
\midrule
DeepSeek-R1-1.5B & 22.00 & 16.67 & 36.80 & 25.16 \\
Alpha-0.1-1.5B & 28.00 & 21.72 & 49.60 & 33.11 \\
Thinkless-1.5B & 22.50 & 16.16 & 33.20 & 23.95 \\
\rowcolor{mycustomcolor}
SmartThinker-1.5B & \textbf{32.00} & \textbf{23.74} & \textbf{52.00} & \textbf{35.91} \\
\midrule
DeepScaleR-1.5B & 31.00 & 22.73 & 48.40 & 34.04 \\
L1-Max-1.5B & 31.50 & 34.37 & 54.00 & 39.96 \\
\rowcolor{mycustomcolor}
SmartThinker-ScaleR & \textbf{33.00} & \textbf{35.83} & \textbf{55.60} & \textbf{41.48} \\
\midrule
DeepSeek-R1-7B & 45.50 & 34.34 & 54.80 & 44.88 \\
Alpha-0.1-7B & 49.00 & 38.89 & 57.60 & 48.50 \\
AdaptThink-7B & 50.00 & 39.90 & 54.00 & 47.97 \\
\rowcolor{mycustomcolor}
SmartThinker-7B & \textbf{54.00} & \textbf{44.44} & \textbf{60.80} & \textbf{53.08} \\
\bottomrule
\end{tabularx}
\end{table}

% \subsubsection{Output Distribution Analysis}
% \paragraph{Output Distribution Analysis.}
% As shown in Figure~\ref{after_train}, SmartThinker yields a flatter output distribution than models using overall length penalty. The proportion of effective steps increases by 15.5\% for low-difficulty questions and by 9.3\% for high-difficulty questions, indicating a reduction in redundant steps and a greater focus on critical reasoning. Moreover, the proportion of the total length attributed to effective steps has increased by approximately 10\%. This improvement indicates that SmartThinker successfully relaxes length constraints for critical steps while imposing stricter length penalty on redundant ones, refining the model’s output distribution and enabling higher-quality reasoning. We present case studies of different baselines in Appendix \ref{case studies} to illustrate SmartThinker’s effectiveness.

\paragraph{In-Domain Reasoning.}
Table~\ref{compact_main_results} summarizes the average in-domain trade-off, with full per-benchmark results reported in Appendix Tables~\ref{math_performance} and~\ref{7B_math_performance}. The central pattern is consistent across all backbone families: SmartThinker moves the model toward a better accuracy-efficiency frontier, improving reasoning accuracy while using substantially fewer tokens. On the DeepSeek-R1-Distill-Qwen-1.5B family, SmartThinker achieves the best average Pass@k and Maj@k among the efficient-reasoning baselines while reducing AvgLen by 37\%. This gives it higher reasoning density than both Alpha-0.1 and Thinkless, showing that the model is not merely learning to truncate its output, but to remove reasoning segments that contribute little to the final answer.

The same conclusion holds in the more demanding DeepScaleR setting, where the base model is already stronger and L1-Max already applies aggressive length control. SmartThinker keeps the accuracy of the strong base model essentially intact, slightly improving average Pass@k from 66.53\% to 66.65\%, while cutting AvgLen from 4988 to 2370 tokens. It also improves over L1-Max under a smaller token budget, yielding the highest density. These results indicate that step-level allocation provides a signal that coarse length control misses: useful derivations should be protected even under strong compression, while low-density continuation can be safely pruned. At 7B scale, SmartThinker again improves both Pass@k and Maj@k over the original backbone while reducing average length by 32.5\%. Taken together, the in-domain results support the main hypothesis of this work: efficient reasoning is best achieved by reallocating tokens within the reasoning trace, rather than by applying uniform pressure to the whole response. Section~\ref{sec:controlled} further confirms that these gains cannot be explained by short-context training, the SFT warm-up initialization, or differences in the training recipe.

\paragraph{Context-Budgeted Reasoning.}
Table~\ref{budget_main_results} tests whether higher-density reasoning translates into higher effective throughput under a fixed context budget. We simulate a harness-style setting in which a model answers a shuffled sequence of AIME24 problems in one growing chat and stops when the cumulative context reaches budget \(B\). The metric is the number of problems solved before the budget is exhausted. Please see Appendix~\ref{context_budget_eval_details} for further experimental details.

At very small budgets, all models are bottlenecked by the first few attempts, so the differences are limited. Once the budget allows multiple full attempts, the benefit of step-level allocation becomes clear. At \(B=16\mathrm{K}\), SmartThinker-7B solves 3.40 problems on average, more than DeepSeek-R1-7B and the efficient-reasoning baselines. At \(B=32\mathrm{K}\), the gap widens: SmartThinker-7B solves 5.60 problems, compared with 4.20 for DeepSeek-R1-7B, 3.60 for Alpha-0.1-7B, and 3.40 for AdaptThink-7B. This pattern shows that SmartThinker uses the same context window to create more useful reasoning opportunities while preserving enough reasoning quality for those additional attempts to become correct answers. This is the operational value of reasoning density in agent harnesses: tokens saved from low-contribution detours are converted into additional useful reasoning opportunities, not merely shorter transcripts.

\paragraph{Out-of-Domain Generalization.}
Table~\ref{ood_main_results} evaluates whether the high-density reasoning policy learned from mathematical training transfers to other reasoning domains. The answer is consistently positive: across three OOD benchmarks, SmartThinker achieves the best average accuracy within each backbone family. The improvements are not marginal. SmartThinker-1.5B raises the average accuracy from 25.16 to 35.91 over the original DeepSeek-R1 backbone and SmartThinker-7B reaches 53.08 average accuracy, outperforming both efficient-reasoning baselines by more than four points.
These results suggest that SCPO does not simply fit math-specific answer formats or benchmark-specific shortcuts. Instead, it learns a more general reasoning behavior: allocating more tokens to steps that are consequential for the final answer while compressing steps that are not. Because this principle is domain-agnostic, the resulting compact reasoning policy remains useful when the task shifts from mathematical derivation to program synthesis, scientific knowledge reasoning, and long-context commonsense inference. The OOD gains therefore strengthen the main claim of SmartThinker: high-density reasoning improves not only efficiency, but also the reliability of reasoning across domains.

\paragraph{Training Overhead.}
The step-contribution estimator is implemented efficiently: all leave-one-step-out sequences are constructed in parallel, dynamically packed across rollouts into a single batch, and scored with one batched prefill-only forward pass using vLLM \cite{kwon2023efficient}, completely avoiding autoregressive decoding (implementation details and full timings are given in Appendix~\ref{app:estimator_cost}). One complete contribution evaluation over eight rollouts of up to 4,096 tokens each takes only 0.60s on the 1.5B model and 4.06s on the 7B model, and end-to-end SCPO training adds less than 8\% per-step overhead over plain GRPO. Since the estimator runs only during training and leaves the decoding path untouched, SmartThinker incurs zero inference-time cost: this one-time training overhead is small and practically acceptable, and is repaid many times over by the sustained inference-efficiency gains reported above.

\begin{figure}[t]
    \begin{center}
		\includegraphics[width=1\linewidth]{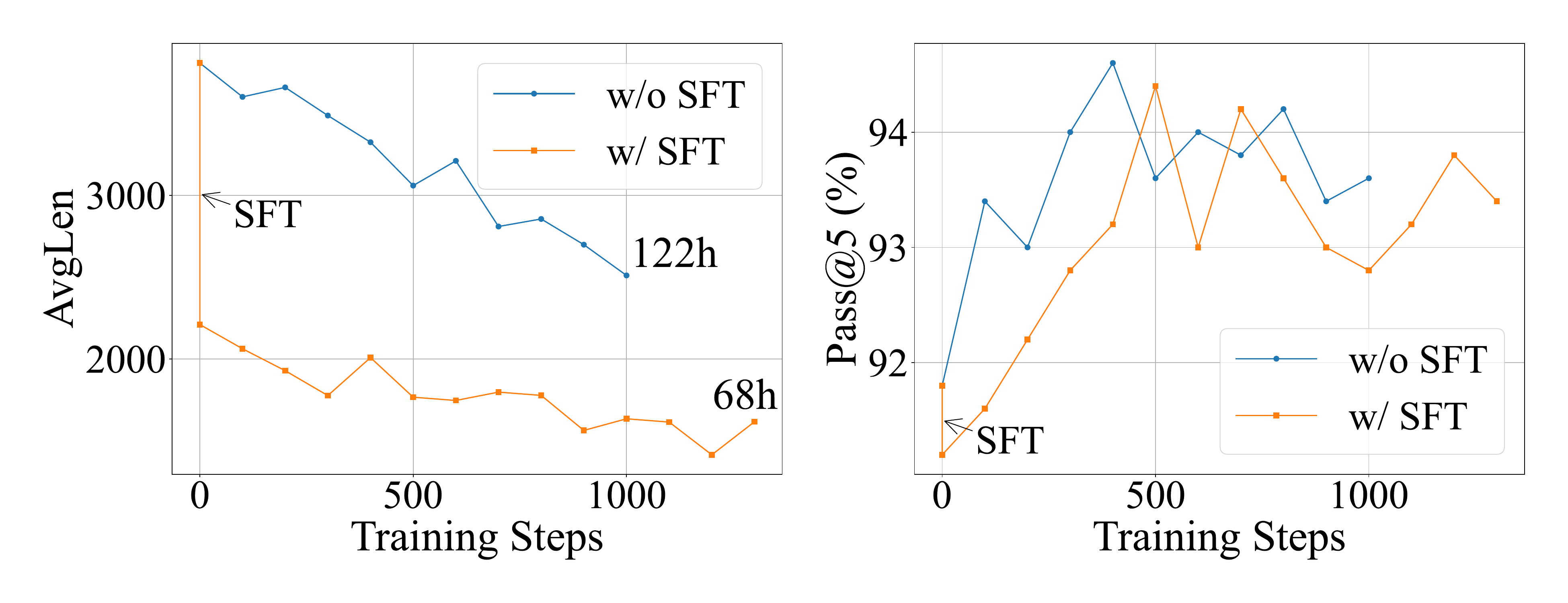}
    \end{center}
    \caption{Pass@5 accuracy and AvgLen at different training steps on the MATH dataset.}
    \label{sft_with_rl}
\end{figure}

\begin{figure*}[t]
    \centering
    % 第二个子图
    \begin{subfigure}[b]{0.32\linewidth}
        \includegraphics[width=\linewidth]{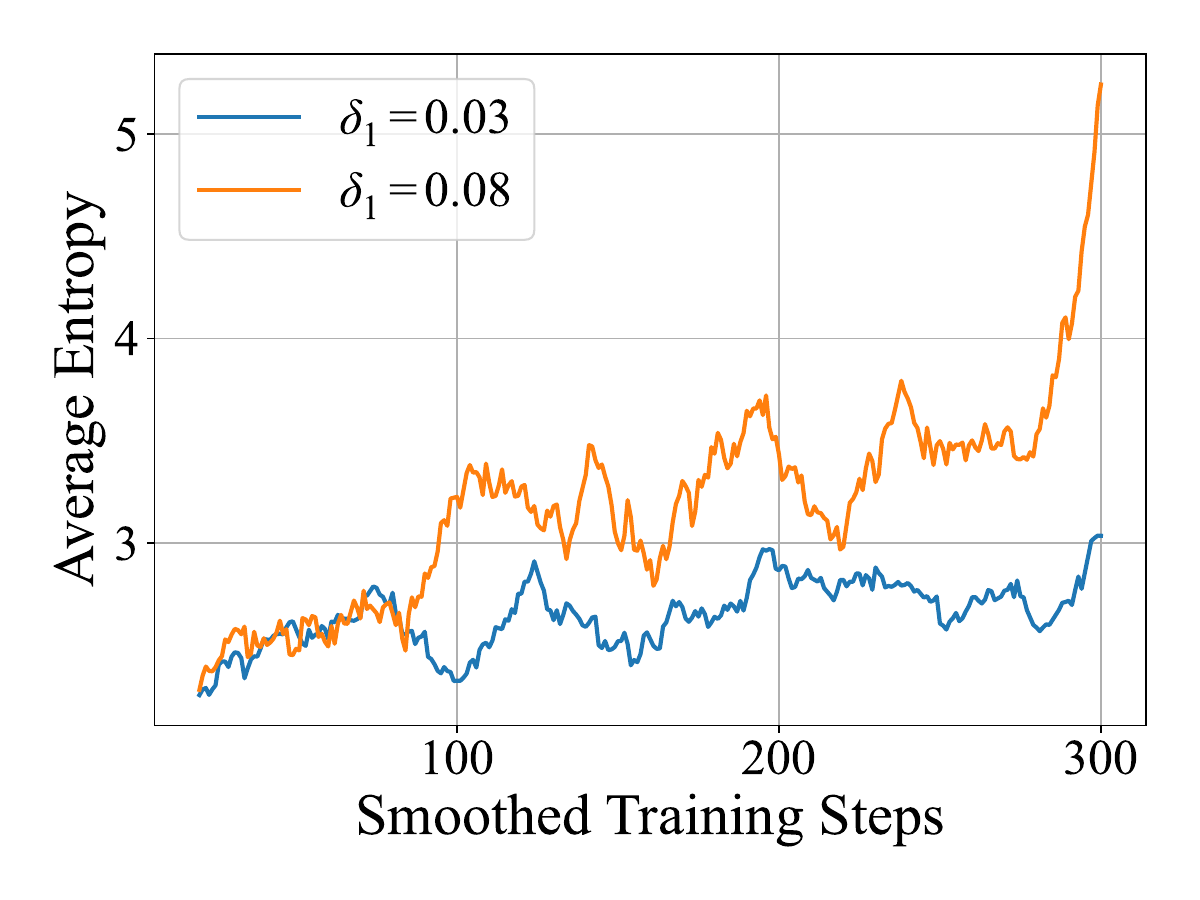}
        \caption{}
        % \caption{Effect of different $\delta_1$ on entropy during training}
        \label{delta_1} 
    \end{subfigure}
    \hfill
    % 第一个子图
    \begin{subfigure}[b]{0.32\linewidth}
        \includegraphics[width=\linewidth]{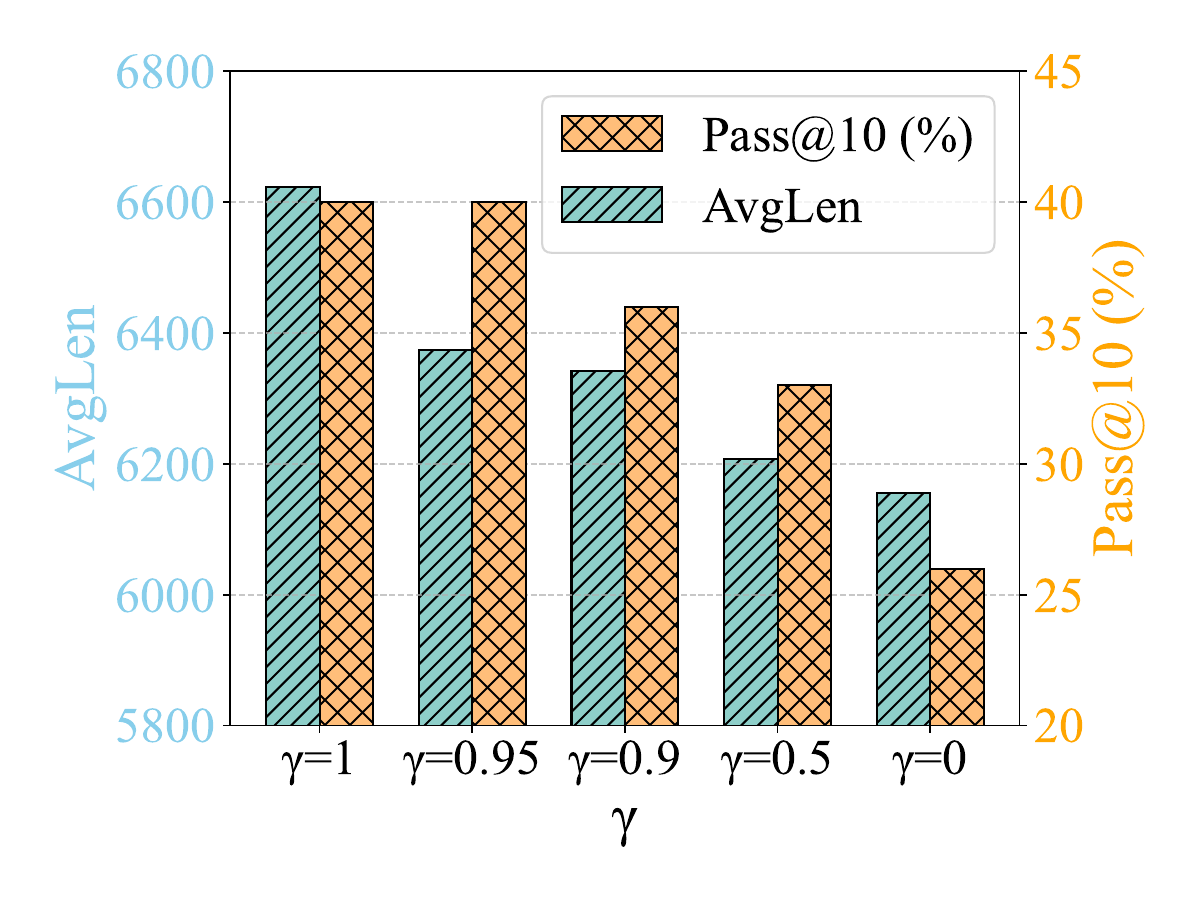}
        \caption{}
        % \caption{Comparison of AvgLen and Pass@10 results on the AIME 2025 dataset under different values of $\gamma$.}
        \label{s-gae}
    \end{subfigure}
    \hfill
    % 第三个子图
    \begin{subfigure}[b]{0.33\linewidth}
        \includegraphics[width=\linewidth]{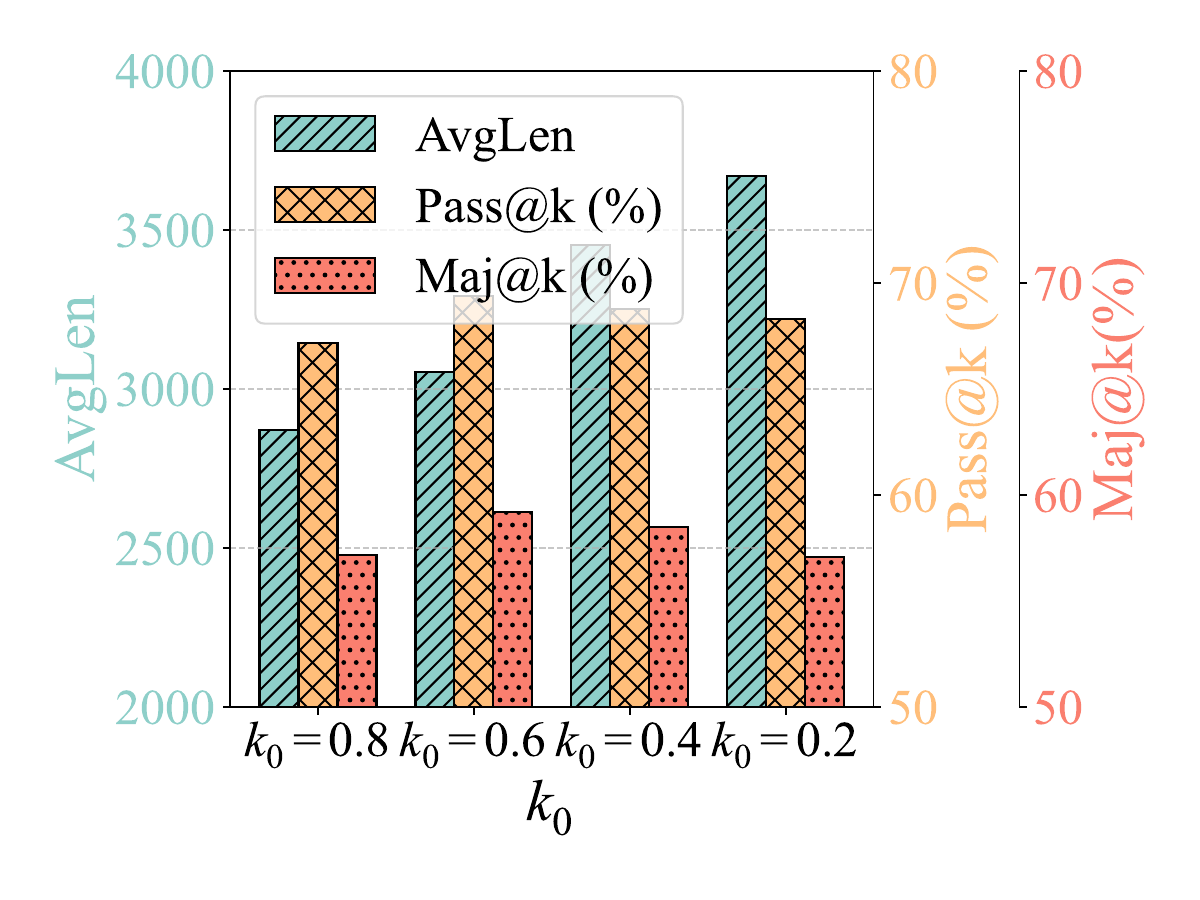}
        \caption{}
        % \caption{Impact of different values of $k_0$ on the model’s reasoning accuracy and efficiency}
        \label{different_k} 
    \end{subfigure}

    \caption{Hyperparameter selection.}
    \label{hyper parameter selection}
\end{figure*}

\section{Ablation Study}
\subsection{Module Ablation Analysis}

\begin{table}[t]
\renewcommand{\arraystretch}{1.05}
\setlength{\tabcolsep}{2pt}
\centering
\footnotesize
\caption{Cumulative module-level ablation on DeepSeek-R1-Distill-Qwen-7B. Results are averaged over the six in-domain benchmarks. Each ``+'' row adds the corresponding component to the previous setting. Dens. denotes average Pass@k per 1K generated tokens.}
\label{module_ablation}
\begin{tabularx}{\columnwidth}{@{}Xccc@{}}
\toprule
Setting & Pass@k & AvgLen & Dens. \\
\midrule
DeepSeek-R1-7B & 71.51 & 4987 & 14.3 \\
+ SFT warm-up & 71.13 & 4001 & 17.8 \\
+ Step contribution estimator & 73.38 & 3673 & 20.0 \\
+ Lightweight safeguard & 74.92 & 3861 & 19.4 \\
+ S-GAE & 74.83 & 3452 & 21.7 \\
\rowcolor{mycustomcolor}
+ Difficulty-adaptive clipping & \textbf{75.26} & \textbf{3367} & \textbf{22.4} \\
\bottomrule
\end{tabularx}
\end{table}

Table~\ref{module_ablation} isolates the contribution of each SmartThinker component on DeepSeek-R1-Distill-Qwen-7B. The SFT warm-up primarily shifts the policy into a shorter-response regime. Although it substantially improves reasoning density, it does not improve accuracy on its own. Figure~\ref{sft_with_rl} further corroborates this interpretation: with SFT warm-up, SCPO attains comparable accuracy at a substantially lower training cost, completing 1,300 RL steps in 68 hours, as opposed to 1,000 steps in 122 hours without SFT. These findings indicate that the warm-up is best viewed as an initialization procedure that reduces rollout cost and better prepares the model for subsequent RL.

The online step contribution estimator produces the first clear accuracy improvement while further reducing response length. This result supports the central design choice of SCPO: reasoning quality improves only when the reward signal can distinguish high-density steps from low-value detours. The lightweight safeguard then raises accuracy at the cost of a small length increase, suggesting that some verification and self-correction steps are worth preserving, especially on difficult problems. Appendix~\ref{app:lightweight_safeguard} further analyzes this behavior and shows that the lightweight safeguard prevents useful reflective steps from being removed too aggressively. This supports a more nuanced view of efficiency: not every verification and reflection step is redundant; the key question is whether the step supports answer-changing reasoning.

S-GAE restores a stronger efficiency gain while maintaining nearly the same accuracy, indicating that step-level credit assignment helps remove the length bias introduced by accumulating rewards over many steps. Finally, difficulty-adaptive clipping yields the best overall trade-off, showing that exploration should be relaxed on hard questions but constrained on easier ones. Appendix~\ref{app:entropy_difficulty} further supports this interpretation by showing that the model's entropy tracks problem difficulty during training. Taken together, the cumulative ablation shows that SmartThinker does not work through indiscriminate shortening. Its advantage comes from progressively turning length control into step-level token allocation, which improves both accuracy and reasoning density.

\subsection{Controlled Comparisons under Identical Warm-up and Recipe}
\label{sec:controlled}

Our gains could in principle come from the warm-up or the 4K recipe rather than SCPO itself. To isolate the RL objective, we retrain every baseline from the same SFT checkpoint with identical data, context, optimization, and training steps; only the RL objective differs. Table~\ref{tab:controlled} shows the pattern: the shortest-answer warm-up compresses every method, but most baselines trade accuracy for brevity, whereas SCPO dominates on both axes. The LAPO pair further shows that adding global length learning on top of this prior hurts accuracy: LAPO trails the released LAPO-I \cite{wu2025lapo} by 2.69 Pass@k points.

\begin{table}[t]
\renewcommand{\arraystretch}{1.0}
\setlength{\tabcolsep}{2pt}
\centering
\footnotesize
\caption{Controlled comparisons under identical SFT warm-up, training data, 4K context, and training steps. Only the RL objective differs across rows; results are averaged over the six in-domain benchmarks. $^{\dagger}$LAPO-I is the publicly released checkpoint trained without our SFT warm-up.}
\label{tab:controlled}
\begin{tabular}{@{}l*{5}{c}@{}}
\toprule
Models & Pass@k & AvgLen. & Red. & Dens. \\
\midrule
SFT+Alpha-0.1-1.5B & 65.47 & 3991 & 30.8 & 16.4 \\
SFT+Thinkless-1.5B & 60.59 & 3897 & 32.4 & 15.6 \\
\rowcolor{mycustomcolor}
SmartThinker-1.5B & \textbf{65.75} & \textbf{3651} & \textbf{36.7} & \textbf{18.0} \\
\midrule
SFT+GRPO-ScaleR & 64.07 & 3030 & 39.3 & 21.1 \\
SFT+L1-Max-1.5B & 63.59 & 2460 & 50.7 & 25.8 \\
\hdashline
LAPO-I-ScaleR$^{\dagger}$ & \textbf{68.94} & 3710 & 25.6 & 18.6 \\
SFT+LAPO-ScaleR & 66.25 & 2896 & 41.9 & 22.9 \\
\rowcolor{mycustomcolor}
SmartThinker-ScaleR & 66.65 & \textbf{2370} & \textbf{52.5} & \textbf{28.1} \\
\midrule
SFT+GRPO-7B & 73.65 & 3687 & 26.1 & 20.0 \\
\rowcolor{mycustomcolor}
SmartThinker-7B & \textbf{75.26} & \textbf{3367} & \textbf{32.5} & \textbf{22.4} \\
\bottomrule
\end{tabular}
\end{table}

Shortest-correct SFT is a strong compression prior: it reduces redundant unrolling and RL exploration cost, but it also narrows the set of correct long-form trajectories. On hard problems that require search, backtracking, or verification, global length penalties or fixed-budget objectives can further suppress key derivations. SCPO recovers accuracy by separating redundant from critical steps with step-level rewards, while its lightweight safeguard and Difficulty-Adaptive Clipping preserve the exploration needed on difficult problems.

\subsection{Entropy-Stable Clipping}

% \begin{figure}[ht]
%     \begin{center}
% 		\includegraphics[width=0.95\linewidth]{picture/s-gae.pdf}
%     \end{center}
%     \caption{Comparison of AvgLen and Pass@10 results on the AIME 2025 dataset under different values of $\gamma$.}
%     \label{s-gae}
% \end{figure}

% \begin{figure}[ht]
%     \begin{center}
% 		\includegraphics[width=0.95\linewidth]{picture/delta1.pdf}
%     \end{center}
%     \caption{ Effect of different $\delta_1$ on entropy during training}
%     \label{delta_1}
% \end{figure}

% \begin{figure}[ht]
%     \begin{center}
% 		\includegraphics[width=0.95\linewidth]{picture/different_k.pdf}
%     \end{center}
%     \caption{ Impact of different values of $k_0$ on the model’s reasoning accuracy and efficiency}
%     \label{different_k}
% \end{figure}

Following DAPO \cite{yu2025dapo}, we set $\delta_2$ to 0.08 and compare the effects of a larger $\delta_1$ with the smaller value used in this work. As shown in Figure~\ref{delta_1}, a larger $\delta_1$ raises entropy but eventually causes entropy collapse, as excessive exploration during moderately complex tasks drives the model off the correct reasoning path. As training continues, this drift escalates, destabilizing training. In contrast, a smaller $\delta_1$ curbs entropy growth, maintaining a healthier balance between exploration and exploitation across problem difficulties.

\subsection{S-GAE Reduces Length Bias}
As shown in Figure~\ref{s-gae}, when $\gamma = 1$, the model tends to generate excessively long sequences. As $\gamma$ decreases, the average sequence length also shortens. However, setting $\gamma$ too low greatly reduces the contribution of later steps to the overall advantage, leading to less accurate advantage estimates and potentially harming model performance.
Balancing these trade-offs, we set $\gamma = 0.95$, which effectively mitigates length bias while maintaining stable performance.

\subsection{Balanced Length Pressure}
Figure~\ref{different_k} studies the effect of the base penalty coefficient $k_0$, which scales both the density-aware coefficient $k_1$ and the difficulty-aware coefficient $k_2$ in the SCPO reward. A large $k_0$ produces shorter responses, but the pressure becomes too aggressive and can remove useful derivations steps along with redundant self-reflection. A small $k_0$ has the opposite limitation: because both $k_1$ and $k_2$ are weakened, the reward has insufficient leverage to prune low-density steps or discourage excessive step counts on easy questions. We therefore set $k_0 = 0.6$, which gives enough pressure to compress low-value reasoning while preserving steps whose contribution justifies their token cost. This supports the central design of SmartThinker: efficiency gains should come from selective token allocation rather than uniform shortening.

\section{Conclusion}
\vspace{-0.2em}

SmartThinker treats efficient reasoning as allocating tokens rather than simply using fewer of them: reasoning density is uneven, with a few steps carrying most of the evidence and many tokens adding little. Using this signal, SmartThinker compresses low-value detours while preserving correctness-critical derivations. This matters especially in agent harnesses, where reasoning shares a bounded context window with tools, memory, and state. Efficient reasoning should therefore be judged by whether the model spends its budget on the steps that matter.

\section{Limitations}
Our method assumes that a reasoning trajectory can be decomposed into discrete steps, which are then used to estimate contribution scores and assign rewards. In this study, we use a fixed delimiter based on \verb|"\n\n"| to define these steps. Although this strategy is simple and consistent with the formatting of many generated reasoning traces, we have not systematically examined how alternative granularities, such as sentence level units, semantic chunks, or fixed token windows, affect training dynamics and final performance. A more adaptive definition of reasoning steps may further improve the stability and generality of step based length control, and we leave this direction for future work.

\section*{Acknowledgments}
This research is supported by the National Key Research and Development Program of China under the grant No.~2023YFF0725003, the National Natural Science Foundation of China under the grant No.~62376129, the Tianjin Science and Technology Plan Project under the grant No.~25ZGZNGX00100, the Hebei Province Science and Technology Plan Project under the grant No.~26241703D, the Tianjin Science and Technology Plan Project under the grant No.~24JCYBJC01950, the Fundamental Research Funds for the Central Universities under the grant No.~63261005, and the Nankai University-AlphaESS Smart Energy Storage Joint Research Center Project.

\bibliography{custom}

@article{guo2025deepseek,
  title={Deepseek-r1: Incentivizing reasoning capability in llms via reinforcement learning},
  author={Guo, Daya and Yang, Dejian and Zhang, Haowei and Song, Junxiao and Zhang, Ruoyu and Xu, Runxin and Zhu, Qihao and Ma, Shirong and Wang, Peiyi and Bi, Xiao and others},
  journal={arXiv preprint arXiv:2501.12948},
  year={2025}
}

@article{jaech2024openai,
  title={Openai o1 system card},
  author={Jaech, Aaron and Kalai, Adam and Lerer, Adam and Richardson, Adam and El-Kishky, Ahmed and Low, Aiden and Helyar, Alec and Madry, Aleksander and Beutel, Alex and Carney, Alex and others},
  journal={arXiv preprint arXiv:2412.16720},
  year={2024}
}

@article{kumar2025overthink,
  title={{OverThink}: Slowdown Attacks on Reasoning {LLMs}},
  author={Kumar, Abhinav and Roh, Jaechul and Naseh, Ali and Karpinska, Marzena and Iyyer, Mohit and Houmansadr, Amir and Bagdasarian, Eugene},
  journal={arXiv preprint arXiv:2502.02542},
  year={2025},
  url={https://arxiv.org/abs/2502.02542}
}

@article{sui2025stop,
  title={Stop overthinking: A survey on efficient reasoning for large language models},
  author={Sui, Yang and Chuang, Yu-Neng and Wang, Guanchu and Zhang, Jiamu and Zhang, Tianyi and Yuan, Jiayi and Liu, Hongyi and Wen, Andrew and Zhong, Shaochen and Chen, Hanjie and others},
  journal={arXiv preprint arXiv:2503.16419},
  year={2025}
}

@article{hou2025thinkprune,
  title={Thinkprune: Pruning long chain-of-thought of llms via reinforcement learning},
  author={Hou, Bairu and Zhang, Yang and Ji, Jiabao and Liu, Yujian and Qian, Kaizhi and Andreas, Jacob and Chang, Shiyu},
  journal={arXiv preprint arXiv:2504.01296},
  year={2025}
}

@article{aggarwal2025l1,
  title={L1: Controlling how long a reasoning model thinks with reinforcement learning},
  author={Aggarwal, Pranjal and Welleck, Sean},
  journal={arXiv preprint arXiv:2503.04697},
  year={2025}
}

@article{arora2025training,
  title={Training Language Models to Reason Efficiently},
  author={Arora, Daman and Zanette, Andrea},
  journal={arXiv preprint arXiv:2502.04463},
  year={2025}
}

@article{qu2025survey,
  title={A survey of efficient reasoning for large reasoning models: Language, multimodality, and beyond},
  author={Qu, Xiaoye and Li, Yafu and Su, Zhaochen and Sun, Weigao and Yan, Jianhao and Liu, Dongrui and Cui, Ganqu and Liu, Daizong and Liang, Shuxian and He, Junxian and others},
  journal={arXiv preprint arXiv:2503.21614},
  year={2025}
}

@article{shao2024deepseekmath,
  title={Deepseekmath: Pushing the limits of mathematical reasoning in open language models},
  author={Shao, Zhihong and Wang, Peiyi and Zhu, Qihao and Xu, Runxin and Song, Junxiao and Bi, Xiao and Zhang, Haowei and Zhang, Mingchuan and Li, YK and Wu, Y and others},
  journal={arXiv preprint arXiv:2402.03300},
  year={2024}
}

@article{hendrycks2021measuring,
  title={Measuring mathematical problem solving with the math dataset},
  author={Hendrycks, Dan and Burns, Collin and Kadavath, Saurav and Arora, Akul and Basart, Steven and Tang, Eric and Song, Dawn and Steinhardt, Jacob},
  journal={arXiv preprint arXiv:2103.03874},
  year={2021}
}

@article{lewkowycz2022solving,
  title={Solving quantitative reasoning problems with language models},
  author={Lewkowycz, Aitor and Andreassen, Anders and Dohan, David and Dyer, Ethan and Michalewski, Henryk and Ramasesh, Vinay and Slone, Ambrose and Anil, Cem and Schlag, Imanol and Gutman-Solo, Theo and others},
  journal={Advances in Neural Information Processing Systems},
  volume={35},
  pages={3843--3857},
  year={2022}
}

@article{he2024olympiadbench,
  title={Olympiadbench: A challenging benchmark for promoting agi with olympiad-level bilingual multimodal scientific problems},
  author={He, Chaoqun and Luo, Renjie and Bai, Yuzhuo and Hu, Shengding and Thai, Zhen Leng and Shen, Junhao and Hu, Jinyi and Han, Xu and Huang, Yujie and Zhang, Yuxiang and others},
  journal={arXiv preprint arXiv:2402.14008},
  year={2024}
}

@misc{veeraboina1983aime,
  title={{AIME} Problem Set 1983--2024},
  author={Veeraboina, Hemish},
  year={2023},
  url={https://www.kaggle.com/datasets/hemishveeraboina/aime-problem-set-1983-2024},
  note={Kaggle dataset}
}

@inproceedings{lightman2023let,
  title={Let's verify step by step},
  author={Lightman, Hunter and Kosaraju, Vineet and Burda, Yuri and Edwards, Harrison and Baker, Bowen and Lee, Teddy and Leike, Jan and Schulman, John and Sutskever, Ilya and Cobbe, Karl},
  booktitle={The Twelfth International Conference on Learning Representations},
  year={2024}
}

@article{fang2025thinkless,
  title={Thinkless: Llm learns when to think},
  author={Fang, Gongfan and Ma, Xinyin and Wang, Xinchao},
  journal={arXiv preprint arXiv:2505.13379},
  year={2025}
}

@article{fu2024efficiently,
  title={Efficiently Serving LLM Reasoning Programs with Certaindex},
  author={Fu, Yichao and Chen, Junda and Zhu, Siqi and Fu, Zheyu and Dai, Zhongdongming and Qiao, Aurick and Zhang, Hao},
  journal={arXiv preprint arXiv:2412.20993},
  year={2024}
}

@article{ye2025limo,
  title={LIMO: Less is More for Reasoning},
  author={Ye, Yixin and Huang, Zhen and Xiao, Yang and Chern, Ethan and Xia, Shijie and Liu, Pengfei},
  journal={arXiv preprint arXiv:2502.03387},
  year={2025}
}

@article{xia2025tokenskip,
  title={Tokenskip: Controllable chain-of-thought compression in llms},
  author={Xia, Heming and Li, Yongqi and Leong, Chak Tou and Wang, Wenjie and Li, Wenjie},
  journal={arXiv preprint arXiv:2502.12067},
  year={2025}
}

@article{han2024token,
  title={Token-budget-aware llm reasoning},
  author={Han, Tingxu and Wang, Zhenting and Fang, Chunrong and Zhao, Shiyu and Ma, Shiqing and Chen, Zhenyu},
  journal={arXiv preprint arXiv:2412.18547},
  year={2024}
}

@article{xu2025chain,
  title={Chain of draft: Thinking faster by writing less},
  author={Xu, Silei and Xie, Wenhao and Zhao, Lingxiao and He, Pengcheng},
  journal={arXiv preprint arXiv:2502.18600},
  year={2025}
}

@article{zhang2025lightthinker,
  title={Lightthinker: Thinking step-by-step compression},
  author={Zhang, Jintian and Zhu, Yuqi and Sun, Mengshu and Luo, Yujie and Qiao, Shuofei and Du, Lun and Zheng, Da and Chen, Huajun and Zhang, Ningyu},
  journal={arXiv preprint arXiv:2502.15589},
  year={2025}
}

@article{yan2025inftythink,
  title={Inftythink: Breaking the length limits of long-context reasoning in large language models},
  author={Yan, Yuchen and Shen, Yongliang and Liu, Yang and Jiang, Jin and Zhang, Mengdi and Shao, Jian and Zhuang, Yueting},
  journal={arXiv preprint arXiv:2503.06692},
  year={2025}
}

@article{zhang2025grpo,
  title={Grpo-lead: A difficulty-aware reinforcement learning approach for concise mathematical reasoning in language models},
  author={Zhang, Jixiao and Zuo, Chunsheng},
  journal={arXiv preprint arXiv:2504.09696},
  year={2025}
}

@article{zhang2025adaptthink,
  title={Adaptthink: Reasoning models can learn when to think},
  author={Zhang, Jiajie and Lin, Nianyi and Hou, Lei and Feng, Ling and Li, Juanzi},
  journal={arXiv preprint arXiv:2505.13417},
  year={2025}
}

@article{lou2025adacot,
  title={AdaCoT: Pareto-Optimal Adaptive Chain-of-Thought Triggering via Reinforcement Learning},
  author={Lou, Chenwei and Sun, Zewei and Liang, Xinnian and Qu, Meng and Shen, Wei and Wang, Wenqi and Li, Yuntao and Yang, Qingping and Wu, Shuangzhi},
  journal={arXiv preprint arXiv:2505.11896},
  year={2025}
}

@article{luo2025ada,
  title={Ada-R1: Hybrid-CoT via Bi-Level Adaptive Reasoning Optimization},
  author={Luo, Haotian and He, Haiying and Wang, Yibo and Yang, Jinluan and Liu, Rui and Tan, Naiqiang and Cao, Xiaochun and Tao, Dacheng and Shen, Li},
  journal={arXiv preprint arXiv:2504.21659},
  year={2025}
}

@article{yu2025dapo,
  title={Dapo: An open-source llm reinforcement learning system at scale},
  author={Yu, Qiying and Zhang, Zheng and Zhu, Ruofei and Yuan, Yufeng and Zuo, Xiaochen and Yue, Yu and Dai, Weinan and Fan, Tiantian and Liu, Gaohong and Liu, Lingjun and others},
  journal={arXiv preprint arXiv:2503.14476},
  year={2025}
}

@misc{luo2025deepscaler,
  title={{DeepScaleR}: Surpassing {O1-Preview} with a {1.5B} Model by Scaling {RL}},
  author={Luo, Michael and Tan, Sijun and Wong, Justin and Shi, Xiaoxiang and Tang, William Y and Roongta, Manan and Cai, Colin and Luo, Jeffrey and Zhang, Tianjun and Li, Li Erran and others},
  year={2025},
  url={https://pretty-radio-b75.notion.site/DeepScaleR-Surpassing-O1-Preview-with-a-1-5B-Model-by-Scaling-RL-19681902c1468005bed8ca303013a4e2},
  note={Project blog post}
}

@article{zhu2025surprising,
  title={The surprising effectiveness of negative reinforcement in LLM reasoning},
  author={Zhu, Xinyu and Xia, Mengzhou and Wei, Zhepei and Chen, Wei-Lin and Chen, Danqi and Meng, Yu},
  journal={arXiv preprint arXiv:2506.01347},
  year={2025}
}

@article{liu2025understanding,
  title={Understanding r1-zero-like training: A critical perspective},
  author={Liu, Zichen and Chen, Changyu and Li, Wenjun and Qi, Penghui and Pang, Tianyu and Du, Chao and Lee, Wee Sun and Lin, Min},
  journal={arXiv preprint arXiv:2503.20783},
  year={2025}
}

@inproceedings{rajbhandari2020zero,
  title={Zero: Memory optimizations toward training trillion parameter models},
  author={Rajbhandari, Samyam and Rasley, Jeff and Ruwase, Olatunji and He, Yuxiong},
  booktitle={SC20: International Conference for High Performance Computing, Networking, Storage and Analysis},
  pages={1--16},
  year={2020},
  organization={IEEE}
}

@article{loshchilov2017decoupled,
  title={Decoupled weight decay regularization},
  author={Loshchilov, Ilya and Hutter, Frank},
  journal={arXiv preprint arXiv:1711.05101},
  year={2017}
}

@article{wu2025agentic,
  title={Agentic Reasoning: Reasoning LLMs with Tools for the Deep Research},
  author={Wu, Junde and Zhu, Jiayuan and Liu, Yuyuan},
  journal={arXiv preprint arXiv:2502.04644},
  year={2025}
}

@article{chen2025learning,
  title={Learning to reason with search for llms via reinforcement learning},
  author={Chen, Mingyang and Li, Tianpeng and Sun, Haoze and Zhou, Yijie and Zhu, Chenzheng and Wang, Haofen and Pan, Jeff Z and Zhang, Wen and Chen, Huajun and Yang, Fan and others},
  journal={arXiv preprint arXiv:2503.19470},
  year={2025}
}

@article{yang2024qwen2,
  title={Qwen2. 5-math technical report: Toward mathematical expert model via self-improvement},
  author={Yang, An and Zhang, Beichen and Hui, Binyuan and Gao, Bofei and Yu, Bowen and Li, Chengpeng and Liu, Dayiheng and Tu, Jianhong and Zhou, Jingren and Lin, Junyang and others},
  journal={arXiv preprint arXiv:2409.12122},
  year={2024}
}

@article{luo2024improve,
  title={Improve mathematical reasoning in language models by automated process supervision},
  author={Luo, Liangchen and Liu, Yinxiao and Liu, Rosanne and Phatale, Samrat and Guo, Meiqi and Lara, Harsh and Li, Yunxuan and Shu, Lei and Zhu, Yun and Meng, Lei and others},
  journal={arXiv preprint arXiv:2406.06592},
  year={2024}
}

@article{jain2024livecodebench,
  title={Livecodebench: Holistic and contamination free evaluation of large language models for code},
  author={Jain, Naman and Han, King and Gu, Alex and Li, Wen-Ding and Yan, Fanjia and Zhang, Tianjun and Wang, Sida and Solar-Lezama, Armando and Sen, Koushik and Stoica, Ion},
  journal={arXiv preprint arXiv:2403.07974},
  year={2024}
}

@article{rein2023gpqa,
  title={{GPQA}: A Graduate-Level Google-Proof {Q\&A} Benchmark},
  author={Rein, David and Hou, Betty Li and Stickland, Asa Cooper and Petty, Jackson and Pang, Richard Yuanzhe and Dirani, Julien and Michael, Julian and Bowman, Samuel R.},
  journal={arXiv preprint arXiv:2311.12022},
  year={2023}
}

@inproceedings{sprague2024musr,
  title={{MuSR}: Testing the Limits of Chain-of-thought with Multistep Soft Reasoning},
  author={Sprague, Zayne and Ye, Xi and Bostrom, Kaj and Chaudhuri, Swarat and Durrett, Greg},
  booktitle={International Conference on Learning Representations},
  year={2024}
}

@article{wu2025lapo,
  title={{LAPO}: Internalizing Reasoning Efficiency via Length-Adaptive Policy Optimization},
  author={Wu, Xingyu and Yan, Yuchen and Lyu, Shangke and Wu, Linjuan and Qiu, Yiwen and Shen, Yongliang and Lu, Weiming and Shao, Jian and Xiao, Jun and Zhuang, Yueting},
  journal={arXiv preprint arXiv:2507.15758},
  year={2025}
}

@article{huang2025hapo,
  title={{HAPO}: Training Language Models to Reason Concisely via History-Aware Policy Optimization},
  author={Huang, Chengyu and Zhang, Zhengxin and Cardie, Claire},
  journal={arXiv preprint arXiv:2505.11225},
  year={2025}
}

@article{liu2025alrs,
  title={Learn to Reason Efficiently with Adaptive Length-based Reward Shaping},
  author={Liu, Wei and Zhou, Ruochen and Deng, Yiyun and Huang, Yuzhen and Liu, Junteng and Deng, Yuntian and Zhang, Yizhe and He, Junxian},
  journal={arXiv preprint arXiv:2505.15612},
  year={2025}
}

@article{tu2025when,
  title={Learning When to Think: Shaping Adaptive Reasoning in {R1}-Style Models via Multi-Stage {RL}},
  author={Tu, Songjun and Lin, Jiahao and Zhang, Qichao and Tian, Xiangyu and Li, Linjing and Lan, Xiangyuan and Zhao, Dongbin},
  journal={arXiv preprint arXiv:2505.10832},
  year={2025}
}

@article{wang2025sas,
  title={Stabilizing Efficient Reasoning with Step-Level Advantage Selection},
  author={Wang, Han and Yu, Xiaodong and Wu, Jialian and Liu, Jiang and Sun, Ximeng and Bansal, Mohit and Liu, Zicheng},
  journal={arXiv preprint arXiv:2604.24003},
  year={2026}
}

@article{li2026stepwise,
  title={Stepwise Penalization for Length-Efficient Chain-of-Thought Reasoning},
  author={Li, Xintong and Li, Sha and Lin, Rongmei and Jin, Hongye and Li, Linwei and Cui, Hejie and Zhang, Sarah and Chang, Chia-Yuan and Cheng, Kewei and Fetahu, Besnik and Nigam, Priyanka and Shang, Jingbo and Yin, Bing},
  journal={arXiv preprint arXiv:2603.00296},
  year={2026}
}

@inproceedings{li2026ssvpo,
  title={{SSVPO}: Effective Step-Level Credit Assignment for {RL} Training of Language Models},
  author={Li, Yugu and Cao, Zehong and Qiao, Jianglin and Hu, Siyi},
  booktitle={International Conference on Learning Representations},
  year={2026},
  url={https://openreview.net/forum?id=g33DGvnHYd}
}

@inproceedings{kwon2023efficient,
  title={Efficient Memory Management for Large Language Model Serving with {PagedAttention}},
  author={Kwon, Woosuk and Li, Zhuohan and Zhuang, Siyuan and Sheng, Ying and Zheng, Lianmin and Yu, Cody Hao and Gonzalez, Joseph and Zhang, Hao and Stoica, Ion},
  booktitle={Proceedings of the ACM SIGOPS 29th Symposium on Operating Systems Principles},
  year={2023}
}

\appendix

\section{Experimental Appendix}

\subsection{Reasoning Density Distribution under Global-Level and Step-Level Training}
\label{sec:Fig1_experiment_details}
\begin{figure}[t]
    \centering
    \begin{subfigure}[b]{1\linewidth}
        \includegraphics[width=\linewidth]{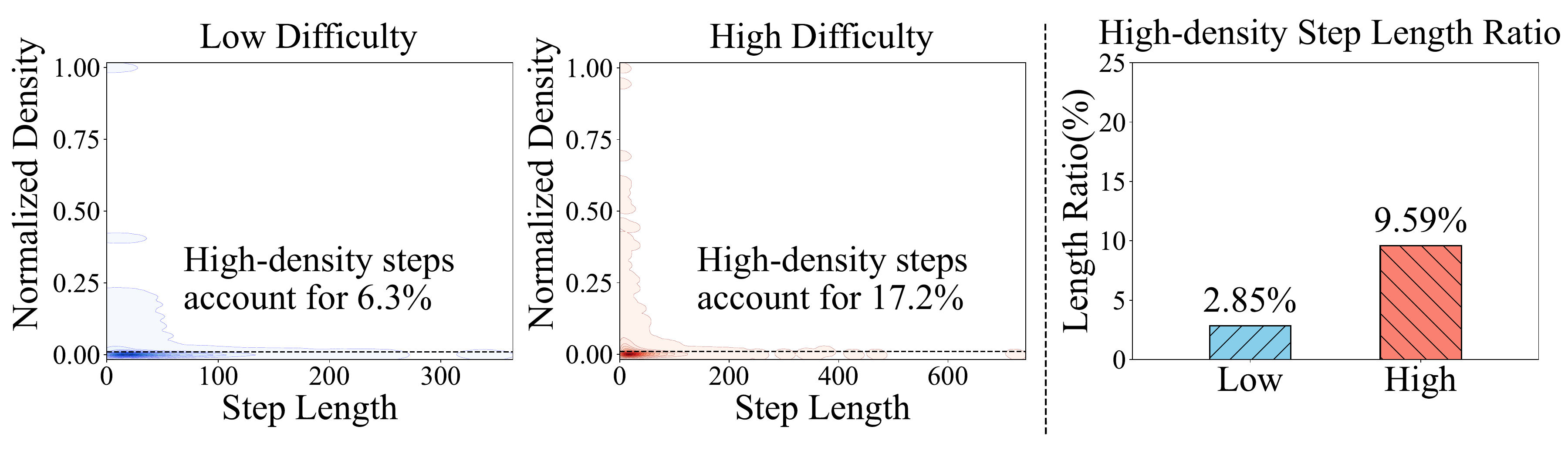}
        \caption{Global-level length-penalty training.}
        \label{before_train}
    \end{subfigure}

    \vspace{1em}

    \begin{subfigure}[b]{1\linewidth}
        \includegraphics[width=\linewidth]{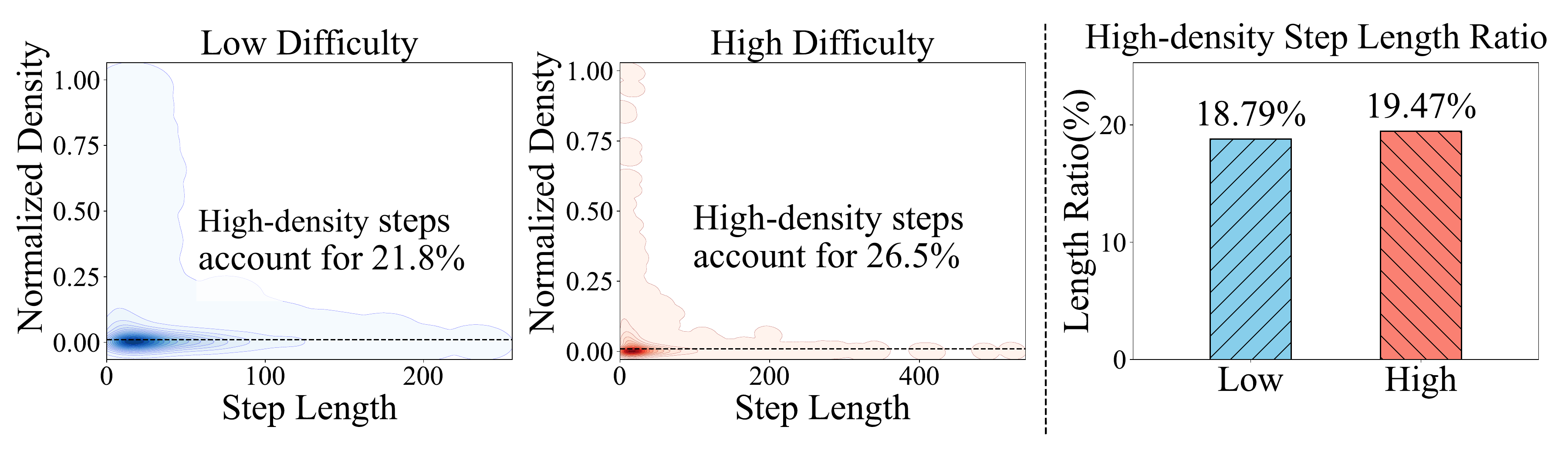}
        \caption{SmartThinker step-level training.}
        \label{after_train}
    \end{subfigure}

    \caption{Reasoning-density distribution under global-level and step-level training. Higher density indicates greater answer contribution per generated token. Compared with global-level length-penalty training, SmartThinker step-level training shifts reasoning toward high-density steps and compresses low-density detours across both easy and hard questions.}
    \label{distribution befor and after train}
\end{figure}

Figure~\ref{distribution befor and after train} compares reasoning-density distributions under global-level and step-level training. Reasoning density measures the estimated answer contribution of a reasoning step per generated token. Under global-level length-penalty training, substantial reasoning remains allocated to low-density steps, indicating redundant token use. Under SmartThinker's step-level training, the distribution shifts toward higher-density reasoning: low-value detours are compressed, while answer-relevant derivations receive more of the reasoning budget. This shows that SmartThinker learns denser reasoning trajectories rather than merely shortening outputs.

Regarding the experimental setup shown in Figure~\ref{distribution befor and after train}, we randomly sample 20 questions from the DeepScaleR-Preview-Dataset and generate 8 responses for each question. Problem difficulty is computed using Eq.\ref{eq:problem_difficulty} in the main text: questions with difficulty scores no greater than 0.5 are categorized as low-difficulty, while those with scores greater than 0.5 are categorized as high-difficulty. We instruct the model to use \verb|\n\n| to separate reasoning steps. For each step $s_{i,j}$, we compute the correct-answer probability with the full trajectory and after removing that step, obtain the raw contribution score using Eq.\ref{eq:raw_contribution}, and divide it by the step length according to Eq.\ref{eq:reasoning_density}. For visualization, we apply the same min--max normalization used in SCPO within each correctness group and aggregate the resulting density scores by training method and problem difficulty.

\subsection{Benchmark Datasets}
\label{benchmark datasets}
In this section, we will describe in detail all the datasets used in our experiments. These datasets are divided into in-domain datasets and out-of-domain datasets.

In-domain datasets are as follows:
\begin{itemize}
    \item 
    AIME24–25 comprises 30 questions from the 2024 and 2025 American Invitational Mathematics Examination, with 15 fill-in-the-blank questions per exam. These questions are more difficult than AMC and span number theory, combinatorics, geometry, and algebra.

    \item
    AMC10/12 consists of 25 multiple-choice questions each for the AMC10 (up to 10th grade) and AMC12 (up to 12th grade). Each competition consists of 25 multiple-choice questions, totaling 975 questions across 39 tests. Questions progress from basic algebra and geometry to introductory probability and counting, covering various tasks for LLM reasoning evaluation.

    \item
    MinervaMATH comprises 12,500 high-school–level contest questions. Each includes detailed solution steps and spans prealgebra through precalculus.

    \item 
    MATH is a challenging dataset used to evaluate models' complex multi-step reasoning abilities in mathematics, featuring problems that require precise logical reasoning where a single error can invalidate an entire solution. It has automatically checkable answers, allowing for outcome supervision without human input, and is employed in comparing outcome and process supervision methods.
    
    \item
    OlympiadBench is an Olympiad-level bilingual multimodal scientific benchmark, featuring 8,476 problems from Olympiad-level mathematics and physics competitions, including the Chinese college entrance exam. Each problem is detailed with expert-level annotations for step-by-step reasoning. Notably, the best-performing model, GPT-4V, attains an average score of 17.97\% on OlympiadBench, with a mere 10.74\% in physics, highlighting the benchmark rigor and the intricacy of physical reasoning.
\end{itemize}

Out-of-domain datasets are as follows:
\begin{itemize}
    \item
    Live-Code-Bench evaluates code generation on recent competitive programming problems collected from platforms such as LeetCode, AtCoder, and Codeforces. Each instance contains a problem statement, input/output examples, and hidden test cases, and models must generate a complete executable solution. We evaluate 200 sampled problems from the release-latest split.

    \item
    GPQA-Diamond is the highest-quality subset of GPQA, a graduate-level multiple-choice benchmark written by domain experts in biology, chemistry, and physics. The questions are designed to be difficult to answer through web search alone and require expert-level scientific reasoning. We evaluate the full 198-question GPQA-Diamond subset.

    \item
    MuSR evaluates multistep soft reasoning over natural-language narratives. We use the MuSR-MurderMysteries subset, where each example requires identifying the correct answer from a long story through chained commonsense and evidential reasoning. We evaluate 250 examples from this subset.
\end{itemize}

\subsection{Baselines}
In this section, we will describe in detail all the baselines used in our experiment.
\begin{itemize}

\item
    % Alpha-0.1 在PPO算法中引入了整体长度惩罚机制，以促使模型生成更简洁的推理过程。该方法在 DeepSeek-R1-Distill-Qwen-1.5B 和 DeepSeek-R1-Distill-Qwen-7B 上进行了全量微调，在显著减少冗余推理的同时保持了较高的正确率。
    Alpha-0.1 introduces a global length penalty mechanism into the PPO algorithm to encourage the model to produce more concise reasoning processes. This approach is fully fine-tuned on both the DeepSeek-R1-Distill-Qwen-1.5B and DeepSeek-R1-Distill-Qwen-7B models, achieving a substantial reduction in redundant reasoning while maintaining high accuracy.

    \item
    % L1-Max 基于Length Controlled Policy Optimization (LCPO)机制对思维链的长度进行精细控制。该方法首先在 DeepScaleR-1.5B-Preview 模型上通过LCPO训练，使模型学会遵循用户设定的 token budget，从而得到基础模型 L1-Exact。随后在此基础上进一步训练出 L1-Max，使其能够生成不超过目标长度的回复。
    L1-Max employs a Length Controlled Policy Optimization (LCPO) mechanism to precisely regulate the length of chain-of-thought reasoning. The method first trains the DeepScaleR-1.5B-Preview model with LCPO, enabling it to follow user-specified token budgets and yielding the base model L1-Exact. Building on this foundation, L1-Max is further trained to generate responses that do not exceed the target length.

    \item
    % Thinkless 采用两阶段训练策略，引导模型根据问题难度动态切换长短推理模式。第一阶段通过有监督微调（SFT）进行预热，使模型掌握长短风格的生成能力；第二阶段引入 Decoupled Group Relative Policy Optimization (DeGRPO) 强化训练，使模型学会在合适的情境下选择输出 <think> 或 <short> 控制符，从而控制推理长度。最终，该方法在 DeepSeek-R1-Distill-Qwen-1.5B 上完成训练，得到 Thinkless 模型。
    Thinkless adopts a two-stage training strategy to guide the model in dynamically switching between short and long reasoning modes based on problem difficulty. In the first stage, supervised fine-tuning (SFT) serves as a warm-up, equipping the model with the ability to generate both concise and elaborate reasoning styles. In the second stage, Decoupled Group Relative Policy Optimization (DeGRPO) is introduced to reinforce training, enabling the model to learn when to output the $\langle\text{short}\rangle$ or $\langle\text{think}\rangle$ control tokens to manage reasoning length. Ultimately, Thinkless is trained on the DeepSeek-R1-Distill-Qwen-1.5B model.

    \item
    % AdaptThink 通过强化学习鼓励模型在可能的情况下直接跳过推理过程，同时结合重要性采样技术，在训练过程中平衡思考样本与非思考样本的比例。最终在 DeepSeek-R1-Distill-Qwen-7B 上训练得到 AdaptThink 模型。
    AdaptThink leverages RL to encourage the model to skip the reasoning process whenever possible, while incorporating importance sampling techniques to balance the proportion of reasoning and non-reasoning samples during training. The final AdaptThink model is trained on DeepSeek-R1-Distill-Qwen-7B.

\end{itemize}

\begin{table*}[t]
    \renewcommand{\arraystretch}{0.73}
    \setlength{\tabcolsep}{4pt}
    \centering
    \caption{Detailed Pass@k, Maj@k, and AvgLen results for 1.5B model families on the six in-domain reasoning benchmarks.}
    \label{math_performance}
    \scalebox{0.78}{
    \begin{tabular}{@{}l*{7}{c}@{}}
    \toprule
    Models & AIME25 & AIME24 & AMC23  & MinervaMATH & MATH & Olympiad-Bench & Avg. \\
    \midrule
    \multicolumn{8}{c}{Pass@k (\%)} \\
    \midrule
    DeepSeek-R1-1.5B & 43.33 & 50.00 & 87.50 & 44.85 & 91.20 & 54.37 & 61.88\\
    Alpha-0.1-1.5B & 40.00 & 63.33 & 90.00 & 49.26 & 92.00 & 56.44 & 65.17\\
    Thinkless-1.5B & 40.00 & 50.00 & 92.50 & 45.59 & 92.00 & \textbf{62.07} & 63.69\\
    \rowcolor{mycustomcolor}
    SmartThinker-1.5B & \textbf{43.33} & \textbf{63.33} & \textbf{95.00} & 45.22 & 91.00 & 56.59 & \textbf{65.75}\\
    \midrule
    DeepScaleR-1.5B-Preview & \textbf{43.33} & 60.00 & \textbf{95.00} & \textbf{48.53} & \textbf{92.60} & \textbf{59.70} & 66.53\\
    L1-Max-1.5B & \textbf{43.33} & 50.00 & \textbf{95.00} & 47.43 & \textbf{92.60} & 57.63 & 64.33\\
    \rowcolor{mycustomcolor}
    SmartThinker-ScaleR & \textbf{43.33} & \textbf{66.67} & 92.50 & 46.32 & 91.80 & 59.26 & \textbf{66.65}\\
    \midrule
    \multicolumn{8}{c}{Maj@k (\%)} \\
    \midrule
    DeepSeek-R1-1.5B & 33.33 & 36.67 & 77.50 & 32.72 & 84.60 & 45.78 & 51.77\\
    Alpha-0.1-1.5B & 30.00 & \textbf{40.00} & 80.00 & \textbf{35.29} & \textbf{86.80} & 46.37 & 53.08\\
    Thinkless-1.5B & 30.00 & 30.00 & 80.00 & \textbf{35.29} & 85.20 & \textbf{51.11} & 51.93\\
    \rowcolor{mycustomcolor}
    SmartThinker-1.5B & \textbf{40.00} & 33.33 & \textbf{82.50} & 33.46 & 83.80 & 49.04 & \textbf{53.69}\\
    \midrule
    DeepScaleR-1.5B-Preview & \textbf{36.67} & \textbf{40.00} & 85.00 & \textbf{38.60} & \textbf{90.00} & \textbf{54.81} & \textbf{57.51}\\
    L1-Max-1.5B & 33.33 & 30.00 & 85.00 & 37.50 & 87.40 & 50.67 & 53.98\\
    \rowcolor{mycustomcolor}
    SmartThinker-ScaleR & 30.00 & \textbf{40.00} & \textbf{90.00} & 33.46 & 88.60 & 50.22 & 55.38\\
    \midrule
    \multicolumn{8}{c}{AvgLen} \\
    \midrule
    DeepSeek-R1-1.5B & 7307 & 7314 & 5335 & 4844 & 3821 & 5985 & 5768\\
    Alpha-0.1-1.5B & 6459 & 6486 & 3771 & 2822 & 2300 & 4596 & 4406\\
    Thinkless-1.5B & 6475 & 6565 & 3888 & 3062 & 1930 & 4249 & 4362\\
    \rowcolor{mycustomcolor}
    SmartThinker-1.5B & \textbf{5653} & \textbf{5310} & \textbf{3105} & \textbf{2495} & \textbf{1643} & \textbf{3700} & \textbf{3651}\\
    \midrule
    DeepScaleR-1.5B-Preview & 6534 & 6833 & 4421 & 4369 & 2888 & 4881 & 4988\\
    L1-Max-1.5B & 3076 & 3255 & 2440 & 3103 & 2221 & 2948 & 2841\\
    \rowcolor{mycustomcolor}
    SmartThinker-ScaleR & \textbf{2748} & \textbf{3004} & \textbf{2217} & \textbf{2072} & \textbf{1885} & \textbf{2291} & \textbf{2370}\\
    \bottomrule
    \end{tabular}
    }
    \end{table*}

    \begin{table*}[t]
        \renewcommand{\arraystretch}{0.73}
        \setlength{\tabcolsep}{4pt}
        \centering
        \caption{Detailed Pass@k, Maj@k, and AvgLen results for 7B models on the six in-domain reasoning benchmarks.}
        \label{7B_math_performance}
        \scalebox{0.8}{
        \begin{tabular}{@{}l*{7}{c}@{}}
        \toprule
        Models & AIME25 & AIME24 & AMC23  & MinervaMATH & MATH & Olympiad-Bench & Avg. \\
        \midrule
        \multicolumn{8}{c}{Pass@k (\%)} \\
        \midrule
        DeepSeek-R1-7B
        & 50.00
        & 73.33
        & 95.00
        & 54.04
        & 95.20
        & 61.48
        & 71.51\\
        Alpha-0.1-7B
        & 60.00
        & 73.33
        & 95.00
        & \textbf{54.78}
        & \textbf{95.60}
        & \textbf{64.30}
        & 73.83
        \\
        AdaptThink-7B
        & 60.00
        & 73.33
        & 95.00
        & 53.31
        & 95.20
        & 63.56
        & 73.40
        \\
        \rowcolor{mycustomcolor}
        SmartThinker-7B
        & \textbf{66.67}
        & \textbf{80.00}
        & 95.00
        & 50.37
        & 95.40
        & 64.15
        & \textbf{75.26}
        \\
        \midrule
        \multicolumn{8}{c}{Maj@k (\%)} \\
        \midrule
        DeepSeek-R1-7B
        & 43.33
        & 56.67
        & 90.00
        & 43.38
        & 91.00
        & 54.07
        & 63.08\\
        Alpha-0.1-7B
        & 43.33
        & 60.00
        & \textbf{95.00}
        & \textbf{44.49}
        & \textbf{92.40}
        & 55.41
        & 65.10
        \\
        AdaptThink-7B
        & 40.00
        & \textbf{70.00}
        & 90.00
        & 42.28
        & 91.80
        & 56.15
        & 65.04
        \\
        \rowcolor{mycustomcolor}
        SmartThinker-7B
        & \textbf{53.33}
        & 63.33
        & 92.50
        & 40.81
        & \textbf{92.40}
        & \textbf{57.19}
        & \textbf{66.59}
        \\
        \midrule
        \multicolumn{8}{c}{AvgLen} \\
        \midrule
        DeepSeek-R1-7B
        & 6820
        & 6791
        & 4600
        & 3589
        & 2815
        & 5308
        & 4987\\
        Alpha-0.1-7B
        & 6265
        & 6161
        & 3466
        & 2322
        & 1887
        & 4267
        & 4061
        \\
        AdaptThink-7B
        & 6332
        & 5962
        & 3185
        & 2002
        & \textbf{1505}
        & 4014
        & 3833
        \\
        \rowcolor{mycustomcolor}
        SmartThinker-7B
        & \textbf{5296}
        & \textbf{5416}
        & \textbf{2777}
        & \textbf{1667}
        & 1570
        & \textbf{3475}
        & \textbf{3367}
        \\
        \bottomrule
        \end{tabular}
        }
        \end{table*}

\subsection{Experimental Setting Details}
\label{setting details}
% In the first stage,  we sample 5 responses per training question and retain only the correct and shortest ones, discarding samples exceeding 4096 tokens. This yields around 2000 samples per model for constructing the synthetic dataset. Sampling sets  temperature to 1 and top-p to 0.95. During SFT, we set the batch size to 1 with gradient accumulation of 8. The learning rate is $1 \times 10^{-5}$, following a cosine schedule with a warm-up ratio of 0.1, and training runs for 3 epochs.
% In the second stage, we train using the TRL framework with a batch size of 8, generating 8 rollouts per question for SCPO reward calculation. We omit the KL penalty, as prior studies show it suppresses exploration \cite{yu2025dapo,liu2025understanding}. The maximum context length is set to 4K tokens during training and extended to 8K for evaluation. Training uses ZeRO Stage 2 \cite{rajbhandari2020zero} on two A6000 GPUs. We set temperature to 1 and top-p to 0.95 during training, and temperature to 0.6 with top-p to 1 during evaluation. To maximize data usage, each batch updates the policy model four times. Hyperparameters are configured as $k_{0} = 0.6$, $\gamma = 0.95$, $\delta_{1} = 0.03$, and $\delta_{2} = 0.08$. Due to hardware limits, For the 1.5B model, we perform full fine-tuning, while for the 7B model, we adopt LoRA \cite{hu2022lora} for lightweight tuning. Optimization is performed with the AdamW \cite{loshchilov2017decoupled} optimizer, using $\beta_1 = 0.9$, $\beta_2 = 0.95$, and a learning rate of $1 \times 10^{-6}$, following a cosine schedule with 60 steps warm-up.
In the first stage, we sample five responses for each training question and retain only those that are both correct and the shortest, discarding any samples exceeding 4096 tokens. This process yields approximately 2,000 samples per model for constructing the synthetic dataset. Sampling uses a temperature of 1 and a top-\(p\) of 0.95. During supervised fine-tuning (SFT), we set the batch size to 1 with a gradient accumulation step of 8. The learning rate is \(1 \times 10^{-5}\), following a cosine schedule with a warm-up ratio of 0.1, and we train for three epochs.

In the second stage, we train the models using the TRL framework with a batch size of 8, generating 8 rollouts per question for SCPO reward computation. We omit the KL penalty, as previous studies show that it suppresses exploration~\cite{yu2025dapo, liu2025understanding}. The maximum context length is set to 4096 tokens during training and extended to 8192 tokens for evaluation. We carry out training with ZeRO Stage 2~\cite{rajbhandari2020zero} across two A6000 GPUs. We set the temperature to 1 and top-\(p\) to 0.95 during training. For in-domain evaluation, we use temperature 0.6. For the OOD evaluation in Table~\ref{ood_main_results}, we use EvalScope with temperature 0.6, top-\(p\) 0.95, and a maximum generation length of 8192 tokens. To maximize data utilization, we update the policy model two times per batch. The hyperparameters are configured as \(k_{0} = 0.6\), \(\gamma = 0.95\), \(\delta_{1} = 0.03\), and \(\delta_{2} = 0.08\). We optimize using the AdamW~\cite{loshchilov2017decoupled} optimizer, with \(\beta_1 = 0.9\), \(\beta_2 = 0.95\), and a learning rate of \(2 \times 10^{-6}\), following a cosine schedule.

\subsection{Detailed Results for the 1.5B Models}
Table~\ref{math_performance} provides the full per-benchmark results for the 1.5B model families. We report the DeepSeek-R1-Distill-Qwen and DeepScaleR groups separately because they start from different base models, so the most meaningful comparisons are within each group. The table includes Pass@k, Maj@k, and AvgLen, allowing us to examine not only whether a method improves accuracy, but also whether it preserves useful reasoning under a shorter token budget.

For the DeepSeek-R1-Distill-Qwen-1.5B family, SmartThinker-1.5B achieves the strongest average Pass@k and Maj@k among the efficient-reasoning methods, while also producing the shortest responses. Compared with Alpha-0.1-1.5B, it improves the average Pass@k from 65.17\% to 65.75\% and Maj@k from 53.08\% to 53.69\%, while reducing AvgLen by 755 tokens. Compared with Thinkless-1.5B, SmartThinker gains 2.06 Pass@k points and 1.76 Maj@k points with 711 fewer tokens. These results suggest that the improvement is not obtained by uniformly truncating reasoning, but by removing low-contribution steps while preserving the derivations that support correct answers.

The DeepScaleR group further tests whether the same idea remains useful when the starting model is already stronger and when the baseline applies aggressive length control. DeepScaleR-1.5B-Preview remains a strong starting point, achieving 66.53\% average Pass@k and the highest average Maj@k in this group. SmartThinker-ScaleR slightly raises average Pass@k to 66.65\% while reducing AvgLen from 4988 to 2370 tokens. Compared with L1-Max-1.5B, it also improves average Pass@k and Maj@k by 2.32 and 1.40 points under a shorter token budget. This supports the central motivation of SCPO: effective compression should depend on step importance rather than a single global length pressure.

\subsection{Detailed Results for the 7B Model}
Table~\ref{7B_math_performance} reports the detailed 7B results across the six in-domain reasoning benchmarks. This setting tests whether the proposed step-level control continues to scale when the base model already has stronger reasoning ability.

SmartThinker-7B obtains the best average Pass@k and Maj@k, improving over DeepSeek-R1-7B by 3.75 and 3.51 points while reducing average response length to 68\% of the original. The gains are especially clear on the most difficult competition-style benchmarks: SmartThinker-7B improves Pass@k by 16.67 points on AIME25 and 6.67 points on AIME24, indicating that its shorter traces do not come at the expense of hard-problem exploration. At the same time, it consistently shortens responses across benchmarks, with the largest reductions on MinervaMATH, AMC23, and Olympiad-Bench.

Compared with Alpha-0.1-7B and AdaptThink-7B, SmartThinker-7B achieves higher average accuracy with fewer tokens. This contrast is informative: global length penalties can shorten responses but may compress important reasoning along with redundant text, while mode-switching methods can reduce cost without directly optimizing the allocation of reasoning effort inside a chain. SmartThinker instead assigns pressure at the step level, which keeps high-contribution derivations available while removing low-value continuation and verification. The 7B results therefore show that SCPO improves the accuracy-efficiency trade-off even when the backbone already has strong reasoning capacity.

\subsection{Controlled Comparison Details}
\label{app:controlled_details}

Table~\ref{tab:controlled_details} reports the full per-benchmark results behind the controlled comparisons in Section~\ref{sec:controlled}. All methods are retrained from the same SFT warm-up checkpoint of each backbone family, with identical training data, 4K training context, optimization configuration, and number of training steps; only the RL objective differs. The publicly released LAPO-I checkpoint (trained without our warm-up) is included as a reference point on the accuracy-efficiency frontier. The table shows that the additional accuracy and compression gains of SmartThinker persist under this strictly controlled setting, so they cannot be attributed to short-context training, initialization, or recipe differences.

\begin{table*}[t]
\renewcommand{\arraystretch}{0.9}
\setlength{\tabcolsep}{3pt}
\centering
\caption{Per-benchmark Pass@k (\%) and average response length of the controlled comparisons described in Section~\ref{sec:controlled}. Each cell pair reports Pass@k and average token count. $^{\dagger}$LAPO-I is the publicly released checkpoint trained without our SFT warm-up and is shown only as a frontier reference.}
\label{tab:controlled_details}
\scalebox{0.83}{
\begin{tabular}{@{}l*{14}{c}@{}}
\toprule
Models & \multicolumn{2}{c}{AIME25} & \multicolumn{2}{c}{AIME24} & \multicolumn{2}{c}{AMC23} & \multicolumn{2}{c}{MinervaMATH} & \multicolumn{2}{c}{MATH} & \multicolumn{2}{c}{Olympiad-Bench} & \multicolumn{2}{c}{Avg.} \\
\cmidrule(lr){2-3} \cmidrule(lr){4-5} \cmidrule(lr){6-7} \cmidrule(lr){8-9} \cmidrule(lr){10-11} \cmidrule(lr){12-13} \cmidrule(lr){14-15}
& P@k & Tok. & P@k & Tok. & P@k & Tok. & P@k & Tok. & P@k & Tok. & P@k & Tok. & P@k & Tok. \\
\midrule
\multicolumn{15}{c}{\textit{DeepSeek-R1-Distill-Qwen-1.5B backbone}} \\
\midrule
SFT+Alpha-0.1-1.5B
& \textbf{43.33} & 5925 & \textbf{63.33} & 5964 & 90.00 & 3405 & \textbf{47.79} & \textbf{2491} & \textbf{92.20} & 1987 & 56.15 & 4172 & 65.47 & 3991 \\
SFT+Thinkless-1.5B
& 36.67 & 5948 & 43.33 & 6073 & 87.50 & 3281 & 45.22 & 2541 & 89.60 & 1711 & \textbf{61.19} & 3825 & 60.59 & 3897 \\
\rowcolor{mycustomcolor}
SmartThinker-1.5B
& \textbf{43.33} & \textbf{5653} & \textbf{63.33} & \textbf{5310} & \textbf{95.00} & \textbf{3105} & 45.22 & 2495 & 91.00 & \textbf{1643} & 56.59 & \textbf{3700} & \textbf{65.75} & \textbf{3651} \\
\midrule
\multicolumn{15}{c}{\textit{DeepScaleR-1.5B-Preview backbone}} \\
\midrule
SFT+GRPO-ScaleR
& 40.00 & 3268 & 53.33 & 3481 & \textbf{95.00} & 2553 & 45.59 & 3296 & 92.00 & 2415 & 58.52 & 3164 & 64.07 & 3030 \\
SFT+L1-Max-1.5B
& 40.00 & 2813 & 53.33 & \textbf{2976} & 92.50 & 2359 & \textbf{47.06} & 2215 & 92.20 & 2001 & 56.44 & 2397 & 63.59 & 2460 \\
SFT+LAPO-ScaleR
& 46.67 & 3769 & 66.67 & 3891 & 90.00 & 2614 & 43.38 & 2458 & 93.00 & 1952 & 57.78 & 2693 & 66.25 & 2896 \\
LAPO-I-ScaleR$^{\dagger}$
& \textbf{50.00} & 5257 & \textbf{73.33} & 5474 & 92.50 & 3295 & 44.49 & 2905 & \textbf{94.20} & 2165 & 59.11 & 3164 & \textbf{68.94} & 3710 \\
\rowcolor{mycustomcolor}
SmartThinker-ScaleR
& 43.33 & \textbf{2748} & 66.67 & 3004 & 92.50 & \textbf{2217} & 46.32 & \textbf{2072} & 91.80 & \textbf{1885} & \textbf{59.26} & \textbf{2291} & 66.65 & \textbf{2370} \\
\midrule
\multicolumn{15}{c}{\textit{DeepSeek-R1-Distill-Qwen-7B backbone}} \\
\midrule
SFT+GRPO-7B
& 63.33 & 6048 & 73.33 & 5923 & \textbf{95.00} & 2987 & \textbf{51.46} & 1809 & 95.20 & 1654 & 63.56 & 3701 & 73.65 & 3687 \\
\rowcolor{mycustomcolor}
SmartThinker-7B
& \textbf{66.67} & \textbf{5296} & \textbf{80.00} & \textbf{5416} & \textbf{95.00} & \textbf{2777} & 50.37 & \textbf{1667} & \textbf{95.40} & \textbf{1570} & \textbf{64.15} & \textbf{3475} & \textbf{75.26} & \textbf{3366} \\
\bottomrule
\end{tabular}
}
\end{table*}

\subsection{Training Cost of the Step-Contribution Estimator}
\label{app:estimator_cost}

This section gives the implementation details and full timings behind the training-overhead paragraph in the main results. Tables~\ref{tab:estimator_cost} and~\ref{tab:training_step_cost} quantify the training overhead of the online step-contribution estimator. For each training step, all leave-one-step-out sequences are constructed in parallel, dynamically packed across rollouts into a single batch, and scored by one batched prefill-only forward pass with vLLM \cite{kwon2023efficient}, so that no autoregressive decoding is involved. Table~\ref{tab:estimator_cost} reports the wall-clock time of one complete contribution evaluation over 8 rollouts of up to 4,096 tokens each, averaged over three independent runs.

\begin{table}[t]
\renewcommand{\arraystretch}{1.05}
\centering
\footnotesize
\caption{Wall-clock time (seconds) of one online step-contribution evaluation over 8 rollouts (up to 4,096 tokens each), measured over three runs.}
\label{tab:estimator_cost}
\begin{tabular}{@{}lcccc@{}}
\toprule
Model & Run 1 & Run 2 & Run 3 & Avg. \\
\midrule
DeepSeek-R1-1.5B & 0.62 & 0.58 & 0.59 & 0.60 \\
DeepSeek-R1-7B & 4.08 & 4.06 & 4.05 & 4.06 \\
\bottomrule
\end{tabular}
\end{table}

\begin{table}[t]
\renewcommand{\arraystretch}{1.05}
\centering
\footnotesize
\caption{Average wall-clock time (seconds) per training step under identical hardware and configuration. SCPO adds less than 8\% end-to-end overhead over plain GRPO on both model scales.}
\label{tab:training_step_cost}
\begin{tabular}{@{}lccc@{}}
\toprule
Model & GRPO & SCPO & SCPO/GRPO \\
\midrule
DeepSeek-R1-1.5B & 9.35 & 9.81 & 104.92\% \\
DeepSeek-R1-7B & 34.92 & 37.66 & 107.80\% \\
\bottomrule
\end{tabular}
\end{table}

Table~\ref{tab:training_step_cost} further compares the end-to-end per-step training time of SCPO and plain GRPO under identical hardware and training configuration. The removal-based estimation adds less than 8\% overhead on both model scales. Since the estimator runs only during training and leaves decoding untouched, SmartThinker incurs no inference-time cost, and this one-time training overhead is repaid by the sustained inference-efficiency gains reported in the main results.

\subsection{Context-Budgeted Evaluation Details}
\label{context_budget_eval_details}
We evaluate context-budgeted reasoning on the 30-problem AIME24 test set. For each run, we shuffle the problems with a fixed seed, append them one by one to a cumulative chat history, and let the model answer the current problem before moving to the next one. The run stops when the accumulated prompt and completion tokens reach the context budget \(B\), or when all 30 problems have been attempted. The main metric in Table~\ref{budget_main_results} is the number of correctly solved problems before stopping.
We use budgets \(B \in \{4096, 8192, 16384, 32768\}\). For each budget, we evaluate five shuffled problem orders and use the same seed set for all models under that budget. Decoding uses temperature 0.6 and top-\(p\) 0.95. At each turn, the maximum generation length is set to the remaining context budget after formatting the current chat prompt. Answers are extracted with the same answer parser and mathematical equivalence checker used in the main evaluation.

\begin{table}[t]
\renewcommand{\arraystretch}{0.9}
\setlength{\tabcolsep}{2pt}
\centering
\caption{Average number of attempted AIME24 problems under each context budget. Attempts include correctly solved, incorrectly answered, and truncated problems before the run stops.}
\label{budget_attempts}
\resizebox{\linewidth}{!}{
\begin{tabular}{lcccc}
\toprule
Models & \(B=4\mathrm{K}\) & \(B=8\mathrm{K}\) & \(B=16\mathrm{K}\) & \(B=32\mathrm{K}\) \\
\midrule
DeepSeek-R1-7B & 1.20 & 1.60 & 4.20 & 16.00 \\
Alpha-0.1-7B & 1.20 & 2.00 & 6.60 & 5.60 \\
AdaptThink-7B & 1.60 & 2.00 & 8.40 & 11.80 \\
\rowcolor{mycustomcolor}
SmartThinker-7B & \textbf{1.60} & \textbf{2.60} & \textbf{12.20} & \textbf{20.80} \\
\bottomrule
\end{tabular}
}
\end{table}

Table~\ref{budget_attempts} shows how many AIME24 problems each model can attempt before exhausting the same cumulative context budget. This auxiliary metric isolates the throughput side of the context-budgeted evaluation. At \(B=4\mathrm{K}\), all models can only make one or two attempts, so the margin is small. As the budget grows, the difference becomes much clearer: SmartThinker-7B attempts 12.20 problems at \(B=16\mathrm{K}\), compared with 4.20 for DeepSeek-R1-7B, 6.60 for Alpha-0.1-7B, and 8.40 for AdaptThink-7B. At \(B=32\mathrm{K}\), SmartThinker-7B reaches 20.80 attempted problems, exceeding the strongest baseline by 4.80 attempts.

Crucially, these additional attempts are not obtained by collapsing reasoning quality. When combined with the solved-problem results in Table~\ref{budget_main_results}, Table~\ref{budget_attempts} shows that SmartThinker converts saved tokens into productive additional opportunities: it attempts the most problems and also solves the most problems at larger budgets. This distinguishes step-level length control from coarse shortening. Rather than simply forcing every response to be brief, SmartThinker removes low-contribution reasoning segments while preserving enough high-value derivation for later attempts to remain useful.

% \subsection{Impact of SFT on RL}
% \label{app:sft_rl}

% Figure~\ref{sft_with_rl} compares SCPO training with and without the short-reasoning SFT warm-up on MATH. In both settings, RL improves Pass@5 while reducing AvgLen, indicating that the step-level optimization signal is effective even without supervised initialization. The main role of SFT is therefore practical: it moves the policy toward a concise but still correct reasoning regime before RL begins, making subsequent rollouts cheaper and training easier to optimize.

% This effect is visible in training cost. Without SFT, the model frequently samples long trajectories and requires 122 hours to complete 1,000 RL steps. With SFT, it completes 1,300 steps in 68 hours while reaching comparable accuracy. Thus, the warm-up is not the core mechanism that creates SmartThinker's final accuracy-efficiency gains; it is an efficient initialization that lets SCPO spend more training budget on learning step-level token allocation rather than repeatedly sampling low-density long traces.

\subsection{Lightweight Safeguard Preserves Useful Reflection}
\label{app:lightweight_safeguard}

\begin{figure}[t]
    \begin{center}
		\includegraphics[width=1\linewidth]{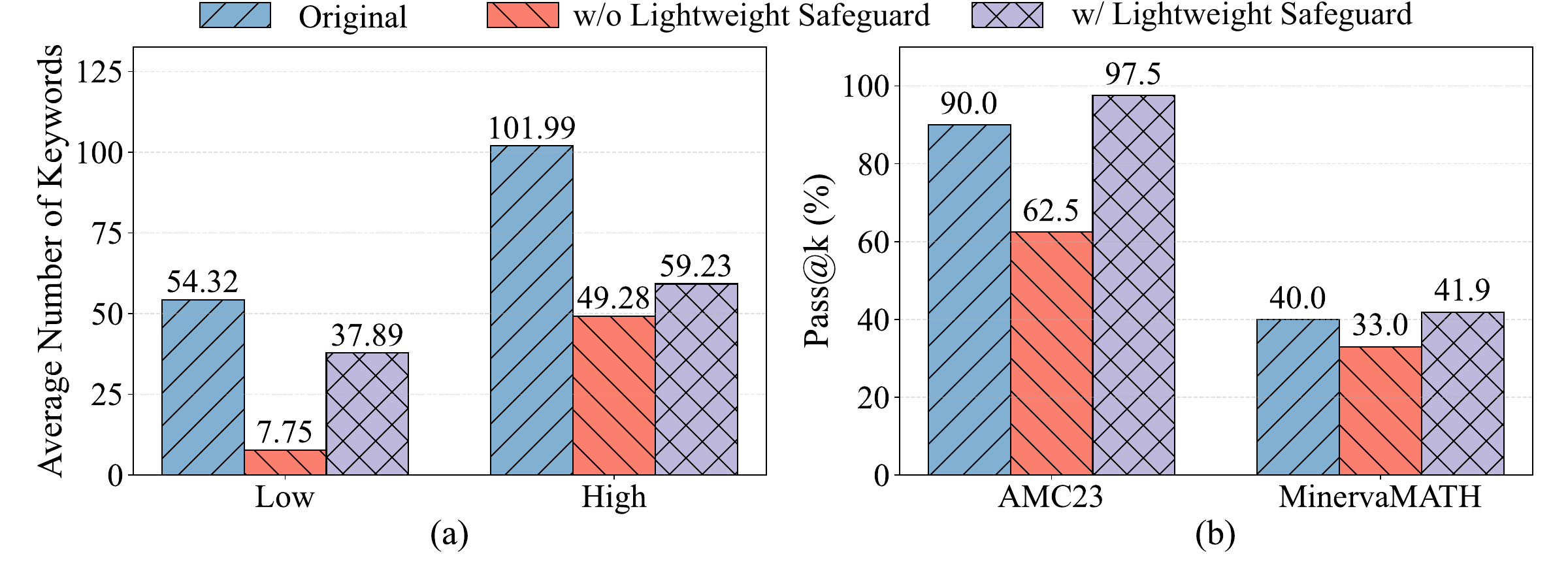}
    \end{center}
    \caption{Effect of lightweight safeguard on reasoning behavior and Pass@k performance across different difficulty levels.}
    \label{additional_scores_all}
\end{figure}

Figure~\ref{additional_scores_all} examines whether the lightweight safeguard preserves useful reflective behavior during compression. Without the safeguard, the model sharply reduces reasoning transition markers, which weakens its ability to revise or verify intermediate conclusions. This failure mode is especially important because reflection is often sparse: many reflective steps are short and may not immediately increase the estimated contribution score, yet they can prevent the model from committing to an early but fragile solution path.

The safeguard addresses this issue by selectively protecting reflective steps instead of relaxing length pressure globally. As a result, the model keeps the verification behavior needed for reliable exploration while still compressing low-value continuation. This explains why the safeguard improves the module-level ablation in Table~\ref{module_ablation}: it does not simply make responses longer, but preserves a small class of steps whose value is easy to underestimate with removal-based scoring alone. Appendix~\ref{reasoning transition words} lists the transition markers used by this auxiliary signal.

\subsection{Difficulty-Adaptive Clipping Tracks Exploration Need}
\label{app:entropy_difficulty}

\begin{figure}[t]
    \begin{center}
		\includegraphics[width=0.83\linewidth]{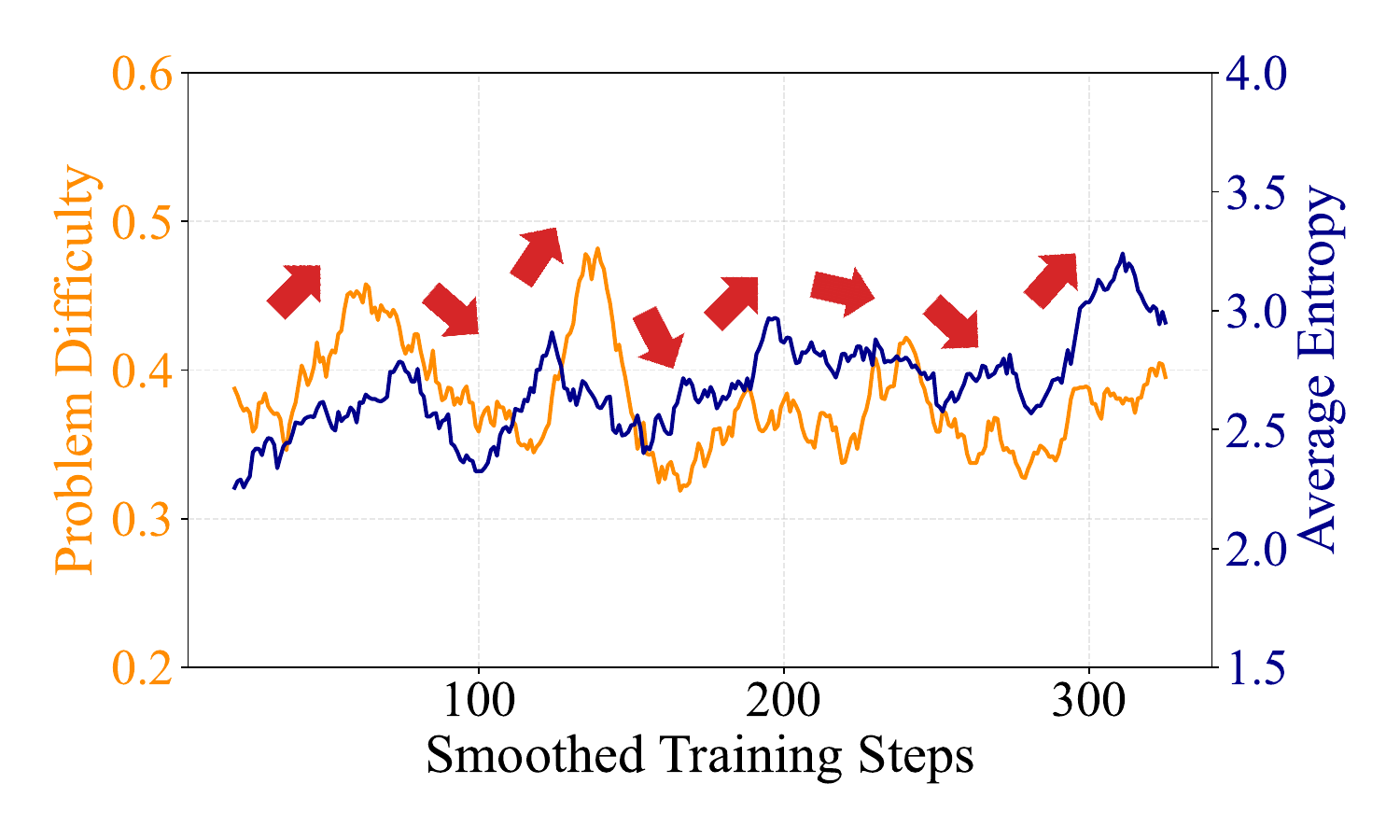}
    \end{center}
    \caption{Entropy trends with problem difficulty during training. Because each sample batch updates the policy model for four iterations, both entropy and problem difficulty are averaged over every four steps.}
    \label{entropy_difficulty}
\end{figure}

Figure~\ref{entropy_difficulty} shows that output entropy rises with problem difficulty and falls when the sampled questions become easier. This pattern is desirable: difficult problems require broader exploration over possible reasoning paths, whereas easy problems benefit from exploiting a stable solution pattern. A fixed clipping range cannot express this distinction and may either under-explore hard problems or destabilize easy ones.

Difficulty-adaptive clipping provides this control by tying the update range to the online difficulty estimate. The resulting entropy trajectory indicates that SmartThinker does not rely on uniformly high exploration to improve accuracy. Instead, it allocates exploration where the rollout set suggests uncertainty and constrains updates where the current policy is already reliable. This behavior explains why the final module in Table~\ref{module_ablation} improves the overall accuracy-efficiency trade-off rather than merely changing response length.

\section{Case Studies}
\label{case studies}

%我们针对不同的 baseline 模型进行了案例分析，以直观展示 SmartThinker 的有效性。如图 7 所示，对于一个可以直接作答的问题，原始模型生成了大量冗长的文字推理，即使已得出正确答案，仍继续展开繁复的推理过程，从而使简单问题变得复杂，并产生了大量多余的 token。Alpha-0.1 在一定程度上缓解了这一冗余现象，但在最终作答时，仍重复了先前的推理内容。Thinkless 虽然能切换至较短的推理模式，但其推理过程仍显复杂。相比之下，SmartThinker 仅通过一句简洁的推理便直接给出最终答案，完全避免了冗余。如图 8 所示，在应对更具挑战性的问题时，原始模型往往先进行大量纯文本推理，随后又多次进行反思与确认，这不仅显著增加了推理所需的计算资源，也未能有效促进最终答案的生成。Alpha-0.1 和 Thinkless 虽能得到正确答案，但在作答后仍存在冗余的自我反思和验证，导致不必要的 token 浪费。相比之下，SmartThinker 更多地通过简洁有力的语句，并辅以数学公式进行推理，仅输出对生成最终答案有实际帮助的数学推导和必要的文字解释，从而实现了更加高效的推理过程。

\begin{figure*}[ht]
    \begin{center}
		\includegraphics[width=0.95\linewidth]{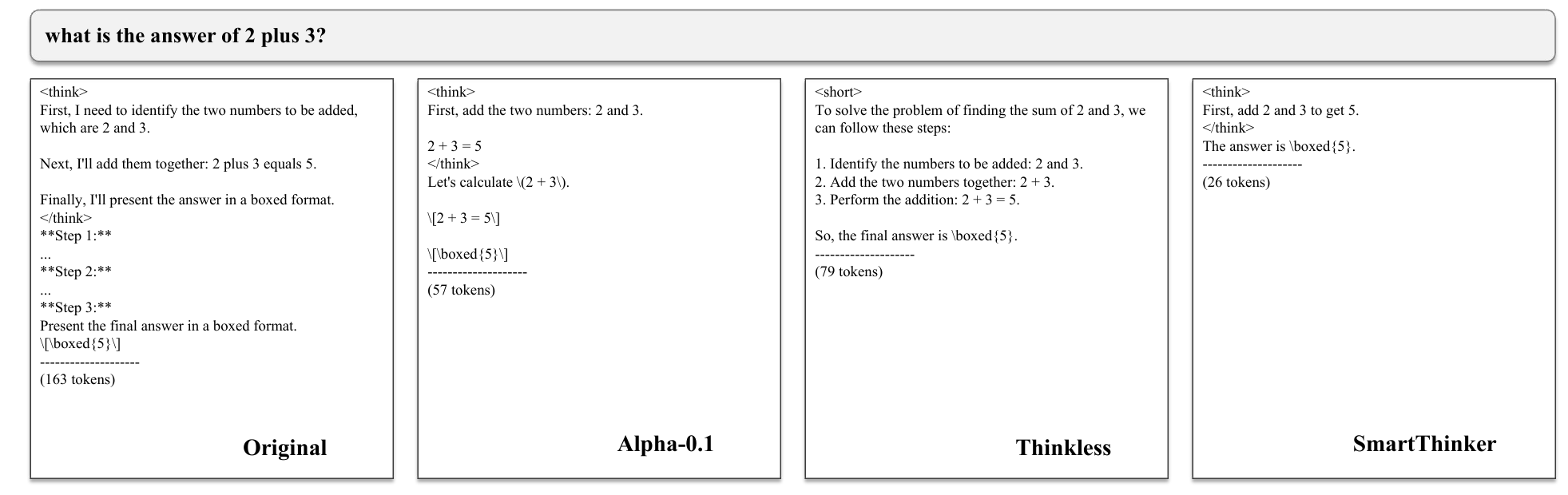}
    \end{center}
    \caption{Examples of model responses to questions with straightforward solutions across different baselines.}
    \label{case_study1}
\end{figure*}

We conduct case studies on different baseline models to provide an intuitive demonstration of SmartThinker’s effectiveness. As shown in Fig. \ref{case_study1}, for a question that allows a direct answer, the original model generates a lengthy textual reasoning and continues to elaborate even after arriving at the correct answer. This unnecessarily complicates simple problems and results in substantial token redundancy. Although Alpha-0.1 mitigates this redundancy to some extent, it still repeats portions of the reasoning process when producing the final answer. Although Thinkless is capable of switching to short-reasoning modes, its reasoning remains relatively complex. In contrast, SmartThinker provides the final answer directly with a single concise line of reasoning, entirely avoiding such redundancy.

\begin{figure*}[ht]
    \begin{center}
		\includegraphics[width=0.95\linewidth]{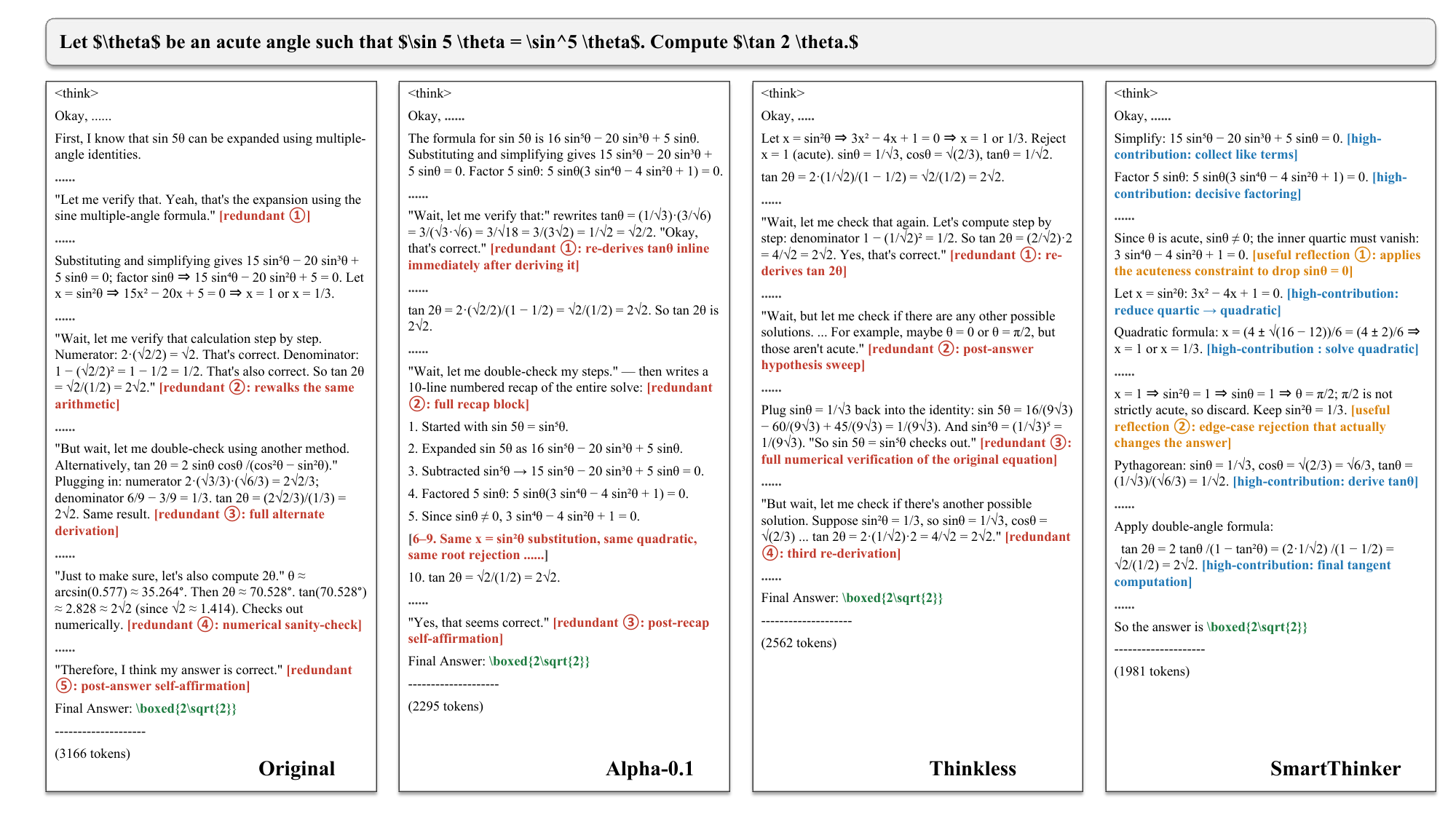}
    \end{center}
    \caption{Examples of model responses to challenging questions across different baselines.}
    \label{case_study3}
\end{figure*}

As shown in Fig. \ref{case_study3}, when faced with more challenging questions, the original model often engages in extensive purely textual reasoning, followed by multiple rounds of reflection and verification. This significantly increases computational resources required for reasoning without contributing meaningfully to producing the final answer. Although Alpha-0.1 and Thinkless are able to arrive at the correct answers, they still perform redundant self-reflection and verification afterward, leading to unnecessary token consumption. In contrast, SmartThinker relies more on concise and effective statements interwoven with mathematical formulas, outputting only the mathematical derivations and essential textual explanations that are genuinely helpful for producing the final answer. This enables a more efficient reasoning process overall.

\section{Theoretical Analysis}
\label{theoretical analysis}
In this section, we provide a theoretical proof that step-level length control can achieve a better trade-off between reasoning quality and computational efficiency compared to global length control. 

Typically, the step-level length control algorithm aims to maximize the following reward:
\begin{align}
R_{\text{step}} = \mathbb{1}(\text{correct}) - \sum_{i=1}^{n} \lambda_{i} \frac{l_{i}}{T},
\end{align}
where $\mathbb{1}(\cdot)$ denotes the indicator function, $\lambda_{i}$ is the length penalty coefficient for the $i$-th step, $l_{i}$ is the length of the $i$-th step, and $T = \sum_{i=1}^{n} l_{i}$ denotes the total length of the response.

The reward for the $i$-th step can therefore be defined as:
\begin{align}
r_{i}^{\text{step}} = m_{i} - \lambda_{i} \frac{l_{i}}{T},
\label{step_reward}
\end{align}
where $m_{i}$ represents the answer contribution of the $i$-th step to the overall correctness of the answer.

In the GRPO algorithm, the advantage for the $i$-th step is defined as:
\begin{align}
A_{i} = \frac{r_{i} - \overline{r}}{\sigma_{r}},
\label{advantage_step}
\end{align}
where $\overline{r}$ and $\sigma_{r}$ denote the mean and standard deviation of the rewards, respectively.

Let the current length of the $i$-th step be $l_{i}$. After an update to the policy gradient, it becomes $l_{i}^{\text{step}}$. For simplicity, we assume that the change in step length is proportional to the policy gradient in reinforcement learning, yielding:
\begin{align}
\Delta l_{i} \propto \mathbb{E} \left[ \nabla_{\theta} \frac{\pi_{\theta}(s_{i} \mid q)}{\pi_{\text{old}}(s_{i} \mid q)} A_{i} \right].
\end{align}
Considering only the variability introduced by advantage estimation, we have:
\begin{align}
l_{i}^{\text{step}} = l_{i} + \eta \cdot A_{i}.
\label{update_length}
\end{align}
Substituting Eq.\ref{step_reward} and \ref{advantage_step} into Eq.\ref{update_length}:
\begin{align}
l_{i}^{\text{step}} = l_{i} \cdot \left(1 - \frac{\alpha \lambda_{i}}{T}\right) + \alpha \cdot \delta_{i},
\label{updated_length}
\end{align}
where $\alpha = \frac{\eta}{\sigma_{r}}$ and $\delta_{i} = m_{i} - \overline{r}$. 

If we focus solely on differences in penalty coefficients, the Eq. \ref{updated_length} can be approximated as:
\begin{align}
l_{i}^{\text{step}} \approx l_{i} \cdot \left(1 - \frac{\alpha \lambda_{i}}{T}\right).
\label{finall_step_length}
\end{align}

Similarly, for the global length control algorithm, after one update, the length of the $i$-th step can be expressed as:
\begin{align}
l_{i}^{\text{global}} \approx l_{i} \cdot \left(1 - \frac{\alpha \lambda}{T}\right).
\label{finall_global_length}
\end{align}

Let $v_{i}$ denote the contribution density of the $i$-th step to producing the correct answer. Then, the total contribution under step-level length control is:
\begin{align}
I_{\text{step}} &= \sum_{i} v_{i} \cdot l_{i}^{\text{step}} \\
&= \sum_{i} v_{i} l_{i} - \frac{\alpha}{T} \sum_{i} v_{i} \lambda_{i} l_{i}.
\end{align}
Similarly, under global length control, the total contribution is:
\begin{align}
I_{\text{global}} &= \sum_{i} v_{i} \cdot l_{i}^{\text{global}} \\
&= \sum_{i} v_{i} l_{i} - \frac{\alpha \lambda}{T} \sum_{i} v_{i} l_{i}.
\end{align}
Define $S = \sum_{i} v_{i} l_{i}$. Then:
\begin{align}
I_{\text{step}} - I_{\text{global}} 
= \frac{\alpha}{T} \left( \lambda S - \sum_{i} v_{i} \lambda_{i} l_{i} \right).
\end{align}

Assuming that both methods yield the same total token budget after the update, we have:
\begin{align}
\sum_{i} l_{i}^{\text{step}} = \sum_{i} l_{i}^{\text{global}},
\label{token_budget1}
\end{align}
and substitute Eq.\ref{finall_step_length} and \ref{finall_global_length} into Eq. \ref{token_budget1}. we can obtain:
\begin{align}
\sum_{i} \lambda_{i} l_{i} = \lambda \sum_{i} l_{i}.
\label{token_budget2}
\end{align}

In the step-level length control algorithm, $\lambda_{i}$ should be inversely proportional to the contribution density of the step. Thus, we define $\lambda_{i} = 1 - v_{i}$, yielding:
\begin{align}
I_{\text{step}} - I_{\text{global}}
= \sum_{i} v_{i}^{2} l_{i} - \frac{S^{2}}{T}.
\end{align}

By the Cauchy–Schwarz inequality:
\begin{align}
\left( \sum_{i} v_{i}^{2} l_{i} \right) \cdot \left( \sum_{i} l_{i} \right) 
\ge \left( \sum_{i} v_{i} l_{i} \right)^{2}.
\end{align}

Therefore, we finally obtain:
\begin{align}
I_{\text{step}} \ge I_{\text{global}}.
\end{align}

This result demonstrates that, under the same token budget, step-level length control preserves a higher total contribution toward producing the correct answer compared to global length penalty methods.

\section{Reasoning Transition Words}
\label{reasoning transition words}

In this section, we list a curated set of commonly used reasoning
transition words and phrases that frequently appear in the LLM's chain-of-thought. We group them by their \emph{reasoning function}
to make them easy to read and to use as a keyword set in our method.

\paragraph{Contrast, Correction, and Turning Points}
These markers indicate a change of direction, contrast, or self-correction
(e.g., catching an error and revising the reasoning).

\emph{but, however, nevertheless, nonetheless, yet, still, though, although, even though, instead, rather, on the other hand, on the contrary, by contrast, in contrast, in reality, in fact, that said, despite this, despite that, conversely. wait, hold on, that is not correct I made a mistake, I miscalculated, I misread the problem, let me correct that, let me fix this, let me reconsider, let me re-evaluate this.}

\paragraph{Checking, Verification, and Self-reflection}

These phrases indicate that the model is explicitly checking or verifying its previous steps, which often correspond to important reflective behavior. 

\emph{let me check, let me check this, let me double-check, let me verify, let me verify the calculation, let me make sure, let us check, let us verify, let us confirm, to verify this, to check this, this does not make sense, this seems wrong, this seems incorrect, I now see that, I see a mistake here, there is an error here, let me try again, let me start over, let me try a different approach, let me try another method, let me take a different approach, instead, I will try.}

\paragraph{Logical Progression and Deduction}
These words and phrases express logical consequence and progression from
one step to the next.

\emph{therefore, thus, hence, so, consequently, as a result, it follows that, this implies that, we deduce that, we conclude that, from this we get.}

\paragraph{Alternative Approaches and Reformulation}
These expressions introduce alternative methods or rephrase the same idea in a different way. 

\emph{alternatively, another way is to, another approach is to, a different approach is to, equivalently, in other words.}

\paragraph{Assumptions, Conditions, and Setup}

These phrases are used when the model sets up assumptions, conditions, or notations, which often structure the reasoning that follows.

\emph{suppose that, assume that, let us suppose, let us assume, let us consider, given that, note that, we note that, observe that, we observe that.}

\paragraph{Meta-reasoning Cues and Heuristics}

These expressions comment on the difficulty, correctness, or reliability
of the current reasoning path.

\emph{this looks promising, this approach may work, this might be wrong, I may have missed something, I need to be careful}

\paragraph{Lightweight Discourse Markers}

These short discourse markers are common in free-form chain-of-thought and often signal local transitions between micro-steps.

\enlargethispage{2\baselineskip}
\emph{okay, alright, well, now, so now, at this point, so far.}

\end{document}